\documentclass[12pt]{article}

\usepackage{algorithm}
\usepackage{algorithmic}
\usepackage{amsmath}  
\usepackage{amsfonts}
\usepackage{amssymb}
\usepackage[dvips]{graphicx}
\usepackage{amsbsy}
\usepackage{amsthm}

\usepackage{latexsym}
\usepackage{subfig}
\usepackage{multicol}
\usepackage{url}
\usepackage{verbatim}
\usepackage[figuresright]{rotating}
\usepackage[varg]{txfonts}
\usepackage{setspace}
\usepackage{tocbibind}
\usepackage{amssymb}
\usepackage[english]{babel}
\usepackage[usenames, dvipsnames]{color}
\usepackage{verbatim}

\usepackage[  dvips, 
                 bookmarks,
                 bookmarksopen = true,
                 bookmarksnumbered = true,
                 breaklinks = true,
                 colorlinks = true,
                 linkcolor = blue,
                 urlcolor  = blue,
                 citecolor = red,
                 anchorcolor = green,
                 hyperindex = true,
                 hyperfigures
                 ]{hyperref}
\newcommand{\secref}[1]{Sect.~\ref{#1}}
\newcommand{\algoref}[1]{Algorithm~\ref{#1}}
\newcommand{\tabref}[1]{Table~\ref{#1}}
\newcommand{\figref}[1]{Fig.~\ref{#1}}

\newcommand{\algoline}[1]{\texttt{\ref{#1}:}}
\renewcommand{\eqref}[1]{Eq.~(\ref{#1})}
\newcommand{\C}[1]{}

\newcommand{\bs}[1]{\boldsymbol{#1}}

\newcommand{\argmax}{\operatornamewithlimits{argmax}}

\newcommand{\x}{\bs{x}}
\newcommand{\X}{\mathbf{X}}

\newcommand{\y}{\bs{y}}
\newcommand{\Y}{\mathbf{Y}}

\newcommand{\al}{\bs{\alpha}}
\newcommand{\alh}{\bs{\widehat{\alpha}}}

\newcommand{\hh}{\bs{\widehat{h}}}

\newcommand{\Hh}{\bs{\widehat{H}}}

\newcommand{\w}{\bs{w}}

\newcommand{\wh}{\bs{\widehat{w}}}

\newcommand{\Phib}{\bs{\varphi}}

\newcommand{\diag}{\mathop{\mathrm{diag}}}
\newcommand{\trace}{\mathop{\mathrm{tr}}}
\newcommand{\median}{\mathop{\mathrm{median}}}

\newcommand{\proposed}{$\ell_1$-LSMI}

\begin{document}

\title{\vspace*{-15mm}
Feature Selection via $\ell_{1}$-Penalized\\
Squared-Loss Mutual Information
}
\author{
Wittawat Jitkrittum\\
Department of Computer Science, Tokyo Institute of Technology\\
2-12-1-W8-74 O-okayama, Meguro-ku, Tokyo 152-8552, Japan\\
\texttt{wittawatj@gmail.com}\\[5mm]
Hirotaka Hachiya\\
Department of Computer Science, Tokyo Institute of Technology\\
2-12-1-W8-74 O-okayama, Meguro-ku, Tokyo 152-8552, Japan\\
\texttt{hacchan@gmail.com}\\[5mm]
Masashi Sugiyama\\
Department of Computer Science, Tokyo Institute of Technology\\
2-12-1-W8-74 O-okayama, Meguro-ku, Tokyo 152-8552, Japan\\
\texttt{sugi@cs.titech.ac.jp}\\
}
\date{}

\maketitle

\begin{abstract}
\noindent
Feature selection is a technique to screen out less important features. 
Many existing supervised feature selection algorithms
use redundancy and relevancy as the main criteria to select features. 
However, \emph{feature interaction}, potentially a key characteristic
in real-world problems, has not received much attention.
As an attempt to take feature interaction into account,
we propose \proposed{}, an $\ell_1$-regularization based algorithm
that maximizes a squared-loss variant of mutual information
between selected features and outputs. 
Numerical results show that \proposed{} performs well in handling redundancy,
detecting non-linear dependency, and considering feature interaction.

\begin{center}
\textbf{Keywords}
\end{center}
feature selection,
$\ell_1$-regularization,
squared-loss mutual information,
density-ratio estimation,
dimensionality reduction
\end{abstract}

\section{Introduction}

Recently, solving real-world complex problems with supervised learning techniques has become more and more common. 
In supervised learning, using all variables as input to a learning algorithm works well when the number of variables is limited. 
However, when the number of variables is large (e.g., gene expression-based patient classification), using all variables in the learning process could lead to overfitting and 
a model interpretability problem \cite{Zhao2010}.

To overcome these problems, feature selection techniques are useful.
Feature selection aims at removing unnecessary variables and retaining only relevant variables
for the target supervised learning task.
Many previous studies \cite{Saeys2007,Suzuki2009} showed that feature selection is useful
in finding relevant variables to gain more insight of the data.
Moreover, the generalization ability of the learned model can be improved through the removal of noisy variables \cite{Peng2005,Langley1994}.

Two conflicting criteria which are commonly used to select features are \emph{relevancy} and \emph{redundancy}.
Features are relevant if they can explain outputs.
Features are redundant if they are similar. 
It is trivial that more features are more likely to explain outputs well.
However, more features are also more prone to be redundant \cite{Peng2005,Zhao2010}.

\emph{Feature interaction} is also another important criterion to consider. 
Feature interaction is a situation in which two or more weak features can explain the output well in the context of each other, even though each of them alone may not be explanatory. 
It is one of the key characteristics in real-world problems.
To detect a group of interacting features, it is necessary to simultaneously consider all features. 
This is because, by definition, considering features individually will not reveal any relevancy to the output. 
Due to this difficulty, feature interaction has not received much attention from the community. 

In this research, instead of focusing on only the relevancy and the redundancy as many previous studies did, we also take into consideration the interaction among features. 
We propose $\ell_1$-LSMI, an $\ell_1$-regularization based algorithm
that maximizes a squared-loss variant of mutual information between selected features and 
outputs. 
We also experimentally compare the proposed method
with several state-of-the-art feature selection algorithms on both artificial and real data.
Numerical results show that $\ell_1$-LSMI performs well in handling redundancy, detecting 
non-linear dependency, and considering feature interaction.

The structure of this paper is as follows.
We formulate our feature selection problem in \secref{sec:formulation}.
Then we describe optimization strategies commonly used in practice in \secref{sec:opt}, as well as several feature quality measures in \secref{sec:measure}. 
We argue that, among the listed strategies, $\ell_1$-regularization based feature weighting is the best choice if we take into account the balance between computation and consideration of features.
As a feature quality measure, we show that squared-loss mutual information (SMI) \cite{Suzuki2009} possesses various desirable properties. 
Based on this argument, in \secref{sec:proposed}, we propose to combine $\ell_1$-regularization and SMI, which we refer to as \proposed{}. 
Experiments on artificial and real data are described in \secref{sec:exp}. 
Finally, we conclude the paper in \secref{sec:conclude}.

\section{Problem Formulation}
\label{sec:formulation}
A formal description of a supervised feature selection problem is as follows. 
Assume we have an input data matrix $\X \in \mathbb{R}^{m \times n}$ and output data vector $\Y \in \mathbb{R}^n$, where $m$ is the number of features and $n$ is the sample size. 
$\X$ and $\Y$ are realizations of the random variable $X=(X_1,\ldots,X_m)$ and $Y$, respectively. Given the desired number of features $k$, supervised feature selection attempts to find a subset of features identified by the set of feature indices $\mathcal{I} \subset \{1,\ldots,m\}$, such that the underlying \emph{feature quality measure} $f$ is maximized. Formally, this can be formulated as an optimization problem as
\begin{equation}
\label{eqn:fs}
\begin{aligned}
& \underset{\mathcal{I} \subset \{1,\ldots,m\}}{\text{maximize}}
& & f(\X_\mathcal{I},\Y) \\
& \text{subject to}
& & |\mathcal{I}| = k, \\
\end{aligned}
\end{equation}
where $|\cdot|$ denotes the set cardinality, and $\X_\mathcal{I}$ denotes the data matrix $\X$ retaining only rows indexed by $\mathcal{I}$.

In general, $f$ can be any function which can quantify the desired characteristics of the selected features.
A popular choice for $f$ is the classification accuracy of a chosen classifier \cite{Kohavi1997}.
While the selected features $\widehat{\mathcal{I}}$ obtained from this approach can yield a good classification accuracy, they are only specifically fit to the predictor in use. 
As a result, an objective interpretation of $\widehat{\mathcal{I}}$ may be difficult \cite{Guyon2003}. 
In this work, we opt to focus on feature selection algorithms which are independent of a predictor for wide applicability.

In practice, searching for a good feature subset to maximize $f$ in a reasonable amount of time can be challenging.
In fact, finding the global optimal feature subset is known to be NP-hard \cite{Weston2003,Masaeli2010a}. 
One way to guarantee that we can obtain the global optimal subset is to perform an exhaustive search over all possible subsets. 
However, since there are $2^m$ possible subsets in total, this approach is impractical for large $m$. 
Clearly, a good \emph{optimization strategy} is needed to efficiently explore the subset space.

As shown above, optimization strategies and feature quality measures are two
important research issues in feature selection.
We describe standard optimization strategies in \secref{sec:opt},
and popular feature quality measures in \secref{sec:measure}. 

\section{Optimization Strategies}
\label{sec:opt}
The optimization strategy defines how to search for a good feature subset.
The complexity of these optimization strategies range, with respect to the number of
features $m$, from linear (feature ranking) to exponential (exhaustive search).
Optimization strategies in general attempt to find features which have high relevancy to the output. 
Higher complexity in some strategies follows from the fact that feature redundancy is also taken into consideration. 
We start the discussion with fast feature ranking technique which does not consider feature redundancy. 

\subsection{Feature Ranking}
Feature ranking is one of the simplest feature optimization strategies. 
Given $m$ features $\{X_1,\ldots,X_m\}$, the feature ranking approach solves the optimization problem of the form
\begin{equation*}
\begin{aligned}
& \underset{\mathcal{I} \subset \{1,\ldots,m\}}{\text{maximize}}
& & \sum_{i \in \mathcal{I}} f(X_i, Y) 
& \text{subject to}
& & |\mathcal{I}| = k. \\
\end{aligned}
\end{equation*}
To solve this problem, we calculate $f(X_i, Y)$ for $i \in \{1,\ldots,m\}$, rank
$X_i$ in the descending order, and then select the top $k$ features. 
The notable feature selection algorithms based on this ranking scheme are
Pearson correlation ranking, SPEC \cite{Zhao2007}, the Laplacian score \cite{He2006},
and the mutual information score \cite{Suzuki2009}.

Although simple and fast, feature ranking considers only the relevancy of features. 
Evaluating each feature individually does not take into account the redundancy among features.
Specifically, if there are many relevant features which are similar in nature, all of them will be ranked top.
This is not desirable since having many similar features is usually as good as having just one.
In other words, $k$ best features are not the best $k$ features \cite{Peng2005}.

\subsection{Sequential Search}
To take feature redundancy into account, the popular sequential search \cite{Kohavi1997,Song2007} can be used.
It comes with two variants: forward and backward search. Forward search works iteratively by maintaining the currently selected features $\mathcal{X}_t$. 
At each step $t$, $\mathcal{X}_t$ is updated with 
\[
\mathcal{X}_t \leftarrow \mathcal{X}_{t-1} \cup \{X_t^*\},
\]
where 
$X_t^* = \argmax_X f(\mathcal{X}_{t-1} \cup \{X\})$ and $\mathcal{X}_0 = \emptyset$. 
The backward search works similarly except that $\mathcal{X}_0$ contains the full feature set. 
At each step, a feature which reduces $f$ the least is removed. 


A potential drawback of the sequential search is its greedy search nature which is independent of $k$. 
That is, the search paths are nested for different values of $k$.
Specifically, it is decremental for the backward search, and incremental for the forward search. 
The result is that, for the backward search, once a feature is removed, it will never be considered again. 
Likewise, for the forward search, once a feature is added, it will never be removed even if it is found to be redundant at latter iterations. 


\subsection{Feature Weighting}

Feature weighting \cite{Tibshirani1996,Zhu2003,Li2006,Liu2009} is an approach which can search for features with a continuous optimization. 
Formally, the feature weighting approach attempts to find
a feature weight vector $\wh \in \mathbb{R}^m$ which is the solution 
of the following optimization problem:
\begin{equation}
\label{eqn:featureweightl1}
\begin{aligned}
& \underset{\w}{\text{maximize}}
& & f(\diag(\w)\X, \Y) \\
& \text{subject to}
& & \|\w\|_1 \leq r,
\end{aligned}
\end{equation}
where $\|\cdot\|_1$ denotes the $\ell_1$-norm,
$\diag(\w)$ is a diagonal matrix with $\w$ placed along its diagonal,
and $r>0$ is the tuning parameter for the radius of the $\ell_1$-ball. It is known
that if $r$ is sufficiently small, then the solution tends to be on a vertex of
the $\ell_1$ simplex, which makes $\wh$ sparse \cite{Tibshirani1996}. 
Features can then be selected according to the non-zero coefficients of the solution $\wh$. 
In fact, observations reveal that the number of non-zero coefficients tends to increase as $r$
increases. 
So, a simple bisection method may be used to search for the value of $r$ which gives $k$ features. 


Unlike the sequential search, the feature weighting approach incorporates $k$ into the problem through $r$ from the beginning. 
So, the solutions for different values of $k$ are not necessarily nested. 
This characteristic is particularly useful when there are multiple optimal feature subsets of different sizes which are disjoint.

\section{Feature Quality Measures}
\label{sec:measure}
In this section, we describe a number of feature quality measures commonly used
in practice. A feature quality measure is a criterion which indicates how good
the selected features are, and is the counterpart of the optimization strategy. 
Here, we focus on predictor-independent criteria.

\subsection{Pearson Correlation}
Pearson correlation is a well-known univariate statistical quantity which can be used to measure a linear dependency between two random variables $X$ and $Y$. It is defined as
\begin{equation}
\rho(X, Y) = \frac{\mathrm{cov}(X,Y)}{\sigma(X) \sigma(Y)},
\label{eqn:pearson}
\end{equation}
where $\mathrm{cov}(X,Y)$ denotes the covariance between $X$ and $Y$, and
$\sigma(X)$ and $\sigma(Y)$ are population standard deviation of $X$ and $Y$,
respectively. 

Although the independence of $X$ and $Y$ implies $\rho=0$, the
converse is not necessarily true since the correlation is capable of detecting
only a linear dependency. 
An example would be a quadratic dependence $Y=X^2$, which gives $\rho=0$ due to the cancellation of
the negatively and the positively correlated components. 

For a feature selection purpose, $|\rho|$ can be used to rank features.
There are many feature selection algorithms based on Pearson correlation \cite{Rodriguez-Lujan2010,Hall2000,Peng2005}.

\subsection{Hilbert-Schmidt Independence Criterion}
The Hilbert-Schmidt independence criterion (HSIC) \cite{Gretton2005} is a
multivariate dependence measure which can detect a non-linear dependency, and
does not require a density estimation. 

The formal definition of HSIC is given as follows. Let $\mathcal{D}_X$ and
$\mathcal{D}_Y$ be the domains of $X$ and $Y$. Define a mapping $\phi(\x) \in
\mathcal{F}$ from all $\x \in \mathcal{D}_X$ to the feature space $\mathcal{F}$
in such a way that the inner product of points in $\mathcal{F}$ is given by a
kernel function $k(\x,\x') = \langle \phi(\x), \phi(\x') \rangle$. 
This can be achieved if $\mathcal{F}$ is a
reproducing kernel Hilbert space on $\mathcal{D}_X$
\cite{AMS:Aronszajn:1950}. Similarly, define another
reproducing kernel Hilbert space.
$\mathcal{G}$ for $\mathcal{D}_Y$ with feature map $\psi$ and kernel $l(\y, \y')
= \langle \psi(\y), \psi(\y') \rangle $. Then, the cross-covariance
operator \cite{Fukumizu2004} associated with the joint probability $p_{xy}$ is a
linear operator $C_{XY}$ defined as
\begin{equation*}
C_{XY} := \mathbb{E}_{\x,\y}[(\phi(\x) 
 - \mu_{\x} ) \otimes (\psi(\y) - \mu_{\y})],
\end{equation*}
where $\otimes$ is the tensor product. HSIC is defined as the squared
Hilbert-Schmidt norm of the cross-covariance operator
\begin{equation*}
\text{HSIC}(p_{xy}, \mathcal{F}, \mathcal{G}) := \|C_{XY}\|^2_{\mathrm{HS}},
\end{equation*}
which could be expressed in terms of kernels \cite{Gretton2005} as 
\begin{align*}
\text{HSIC}(p_{xy}, \mathcal{F}, \mathcal{G}) =& 
 \mathbb{E}_{\x,\x',\y,\y'}[k(\x,\x')l(\y,\y')] \\ 
 & + \mathbb{E}_{\x,\x'}[k(\x,\x')] \mathbb{E}_{\y,\y'}[l(\y,\y')] \\
 & - 2\mathbb{E}_{\x,\y}[\mathbb{E}_{\x'}[k(\x,\x')] \mathbb{E}_{\y'}[l(\y,\y')]
 ].
\end{align*}
$\mathbb{E}_{\x,\x',\y,\y'}[k(\x,\x')l(\y,\y')]$ is the expectation over
independent pairs $(\x,\y)$ and $(\x', \y')$ drawn from $p_{xy}$. Given an
i.i.d.~paired sample $\mathcal{S} = \{(\x_i, \y_i)\}_{i=1}^n$, an empirical estimator  of HSIC is given by
\begin{align}
\text{HSIC}(\mathcal{S}, \mathcal{F}, \mathcal{G}) = \frac{1}{(n-1)^2}
\trace(KHLH),
\label{eqn:hsic_biased} 
\end{align}
where $K,L,H \in \mathbb{R}^{n \times n}, (K)_{i,j} := k(\x_i, \x_j)$, 
$(L)_{i,j}:=l(\y_i, \y_j)$, and $H := I_n - \bs{1}\bs{1}^T/n$ (centering matrix).
It was also shown that, if $k$ and $l$ are universal kernels (e.g., Gaussian kernels)
\cite{JMLR:Steinwart:2001}, then
$\text{HSIC}(p_{xy}, \mathcal{F}, \mathcal{G}) = 0$ if and only if $X$ and $Y$ are independent. So,
HSIC can also be used as a dependence measure. 

In spite of the strong theoretical properties of HSIC,
there is no known objective criterion for model selection of the kernel functions $k$ and $l$. 
A popular heuristic choice is to use a Gaussian kernel with its width set to the median of the pairwise distance of the data points \cite{Scholkopf2002}.

\subsection{Mutual Information}
In information theory, mutual information \cite{Cover1991} is an important
quantity which can be used to detect a general non-linear dependency between two
random variables. 
It has been widely used as the criterion for feature selection \cite{Peng2005,Suzuki2008,Rodriguez-Lujan2010} as well as feature extraction \cite{Torkkola2003}.
Mutual information is defined as
\begin{equation}
I(X,Y) :=  \iint \log\left( 
 \frac{p_{xy}(\x, \y)}{ p_x(\x) p_y(\y)  } 
 \right) p_{xy}(\x, \y) \, \mathrm{d}\x \mathrm{d}\y, 
 \label{eqn:mi}
\end{equation}
which is the Kullback-Leibler divergence \cite{Kullback1951} from
$p_{xy}(\x,\y)$ to $p_x(\x) p_y(\y)$. Mutual information is a measure of dependence in the sense that it is
always non-negative, symmetric ($I(X,Y) = I(Y,X)$), and vanishes if and only if
$X$ and $Y$ are independent, i.e., $p_{xy}(\x,\y) = p_x(\x) p_y(\y)$.

Even though mutual information is a powerful multivariate measure, accurate estimation of the densities $p_{xy}, p_{x}$ and $p_y$ is difficult in high-dimensional case. 
A recent approach which avoids taking the ratio of estimated densities by directly modeling the density ratio $\frac{p_{xy}(\x, \y)}{ p_x(\x)p_y(\y) }$ is Maximum Likelihood Mutual Information (MLMI) \cite{Suzuki2008}. 
Although MLMI was demonstrated to be accurate,
its estimation is computationally rather expensive
due to the existence of the logarithm function.

\subsection{Squared-loss Mutual Information}
Another mutual information variant which has received much attention recently is
Squared-loss Mutual Information
(SMI) \cite{Suzuki2009,Suzuki2010,Hachiya2010,Suzuki2011} defined as
\begin{equation}
I_s(X,Y) :=  \frac{1}{2} \iint \left( 
 \frac{p_{xy}(\x, \y)}{ p_x(\x) p_y(\y) } - 1
 \right)^2 p_x(\x) p_y(\y) \, \mathrm{d}\x \mathrm{d}\y. 
 \label{eqn:smi}
\end{equation}
SMI is based on the $f$-divergence \cite{Ali1966,Csisz'ar1967} with a squared loss
(also known as the Pearson divergence, \cite{Liese2006}), as opposed to the ordinary
mutual information which is based on the $f$-divergence with a log loss (Kullback-Leibler
divergence, \cite{Kullback1951}). Note that $I_s(X,Y)=I_s(Y, X)$, $I_s(X,Y) \geq 0$,
and $I_s(X,Y) = 0$ if and only if $p_{xy}(\x, \y) = p_x(\x)p_y(\y)$, just like the ordinary mutual information.
Therefore, SMI can also be used as a measure of dependence between
$X$ and $Y$.

 SMI can be estimated by directly modeling the ratio $g^*(\x,\y) =
\frac{p_{xy}(\x, \y)}{ p_x(\x) p_y(\y) }$ itself without going through the 
estimation of the densities. The goal is to find a density ratio estimate
$\widehat{g}(\x,\y)$ which is as close to the true density ratio $g^*(\x,\y)$
as possible. Here, the estimation can be formulated as a least-squares problem. 
That is, to find $\widehat{g}(\x,\y)$ such that its expected squared difference from $g^*(\x,\y)$
is minimized:
\begin{align}
\label{eqn:smiloss1}
 & \min_g \frac{1}{2} \iint \left( g(\x, \y) - g^*(\x,
\y) \right)^2 p_x(\x) p_y(\y) \, \mathrm{d}\x \mathrm{d}\y .
\end{align} 
Since finding $g$ over all measurable functions is not
tractable \cite{Suzuki2010}, the model $g$ is restricted to be in a linear
subspace $\mathcal{G}$ defined as
\begin{equation*}
\mathcal{G} := \{\al^T \Phib(\x,\y) \,|\, \al = (\alpha_1, \ldots, \alpha_b)^T
\in \mathbb{R}^b \},
\end{equation*}
where $\al$ is the model parameter to be learned, and $\Phib(\x,\y) =
(\varphi_1(\x,\y),\ldots,\varphi_b(\x,\y))^T$ is a basis function vector such
that $\forall l, \varphi_l(\x,\y) \geq 0$. The basis also admits kernel functions
which depend on samples. 

With $\mathcal{G}$, finding $\widehat{g}$ amounts to finding the optimal
$\al$. By using an empirical approximation,  \eqref{eqn:smiloss1} can be written as
\begin{equation}
\min_{\al \in \mathbb{R}^b} \frac{1}{2} \al^T \Hh \al - \hh^T \al +
\frac{\lambda}{2} \al^T \al, 
\label{eqn:smiloss2}
\end{equation}
where the term $\frac{\lambda}{2} \al^T \al$ with a regularization parameter $\lambda >
0$ is included for a regularization purpose, and
\begin{align*}
\Hh  &:=   \frac{1}{n^2} \sum_{i=1}^n \sum_{j=1}^n \Phib(\x_i, \y_j) 
 \Phib(\x_i, \y_j)^T, 
 \\ 
 \hh  &:=   \frac{1}{n} \sum_{i=1}^n \Phib(\x_i, \y_i).
\end{align*}
 By differentiating \eqref{eqn:smiloss2} with respect to $\al$ and equating it
 to zero, the solution $\alh$ can be computed analytically as
 \begin{equation*}
 \alh =  \left(\Hh + \lambda \bs{I} \right)^{-1} \hh,
 \end{equation*}
where $\bs{I}$ denotes the identity matrix.
Finally, using $\alh$, SMI in \eqref{eqn:smi} can be estimated as
\begin{equation}
\widehat{I}_s  =  \frac{1}{2}\hh^T \alh - \frac{1}{2}. 
\label{eqn:lsmi}
\end{equation}
The estimator in \eqref{eqn:lsmi} is called Least-Squares Mutual Information
(LSMI).

LSMI possesses many good properties \cite{Suzuki2010}. 
For example, it has an optimal convergence rate in $n$ under non-parametric 
setup. 
Also, LSMI is equipped with a model selection criterion for
determining $\Phib$ and $\lambda$. 
Model selection by $K$-fold cross validation  is described as follows. 
First, randomly split samples $\{(\x_i, \y_i)\}_{i=1}^n$  into (roughly) equal $K$
disjoint subsets $\{\mathcal{S}_k\}_{k=1}^K$. 
An estimator $ \alh_{\mathcal{S}_{-k}}$ is then obtained using $\mathcal{S}_{-k} := \{\mathcal{S}_j\}_{j \neq k}$.
Finally, the approximation error for the held-out samples $\mathcal{S}_k$ is computed. 
The procedure is repeated $K$ times, and $(\Phib, \lambda)$ which minimizes the mean $\widehat{J}^{(K-CV)}$ is chosen:
\begin{equation*}
 \widehat{J}^{(K-CV)} := \frac{1}{K} \sum_{k=1}^K \left( \frac{1}{2}
 \alh^T_{\mathcal{S}_{-k}} \Hh_{\mathcal{S}_{k}} \alh_{\mathcal{S}_{-k}} -
 \hh_{\mathcal{S}_{k}}^T \alh_{\mathcal{S}_{-k}} \right).
\end{equation*}

\section{Proposed Method}
\label{sec:proposed}
In this section, we describe our proposed method.

\subsection{Motivations}
As mentioned previously, there are a number of factors which cause the difficulty of feature selection, i.e., non-linear dependency, feature interaction, and feature redundancy.
Although existing combinations of optimization strategies and measures can handle these problems, the trade-off of the computational complexity and the obtained abilities to deal with such issues is not well balanced. 

A summary of properties of common optimization strategies is shown in \tabref{tab:f_opt}. 
Ranking is very fast since it completely disregards feature redundancy and feature interaction, and focuses on only feature relevancy. 
Forward search improves this by maintaining a set of selected features, and greedily adding each feature to the set. 
This allows the forward search to deal with feature redundancy by not adding a redundant feature to the set.
Nevertheless, feature interaction cannot be detected since features are not considered in the presence of each other.
This is why backward search comes to play by starting from the full feature set and iteratively removing a feature instead.
Although this scheme has a potential to detect interacting features, the complexity goes from $O(m)$ to $O(m^2)$
which could be problematic when the number of features, $m$, is large. Considering all strategies, an $\ell_1$-based approach seems to be the optimal choice here.
It offers a continuous optimization which is usually easier than a discrete optimization. Also, since all features are
considered simultaneously by optimizing their weights, it can take into account feature redundancy and feature interaction.

A summary of properties of feature quality measures is shown in \tabref{tab:f_measure}.
PC is very efficient to compute. However, only linear dependency can be identified. 
HSIC can reveal a non-linear dependency. Nonetheless, it is unclear how to objectively choose the right kernel function.
MI is another measure that is capable of detecting a nonlinear dependency but the existence of $\log$ causes computational inefficiency. It can be seen that SMI has balanced properties here. Not only is it able to capture a non-linear dependency,
using a squared loss instead of a log loss also permits its estimator to have an analytic form, which can be efficiently computed. 
 
\tabref{tab:opt_measure} shows the combinations of optimization strategies and feature quality measures. 
Many of them have already been proposed in the past. Exhaustive search is marked impractical since it is computationally intractable.
PC is a univariate measure which considers one feature at a time. 
Combining it with a feature-set optimization strategy (i.e., forward, backward search, $\ell_1$ approach) would degenerate
back to a ranking approach. Hence, the combinations are marked unreasonable. 

It can be seen that the feature weighting with $\ell_1$-regularization is the best among the optimization strategies. 
Also, SMI has the best balance among the listed feature quality measures. We therefore propose to combine $\ell_1$-regularized feature weighting with SMI, which we call \proposed{}.

\begin{table}[t]
\centering
\caption{Summary of properties of optimization strategies.
``disc.'' and ``cont.'' denote ``discrete'' and ``continuous'', respectively.}
\label{tab:f_opt}
\begin{tabular}{@{}c|ccccc@{}}
\hline 
& {Ranking} & {Forward} & {Backward} & {Exhaustive} & {$\ell_1$}\\
\hline \hline 
Optimization & disc. & disc. & disc. & disc. & cont. \\
Complexity
& $m$ & $m$ & $m^2$ & $2^m$ & $m$ \\
Redundancy
& $\times$ & $\triangle$ & $\bigcirc$ &$\circledcirc$ & $\bigcirc$ \\
Interaction
& $\times$ & $\times$& $\bigcirc$ & $\circledcirc$& $\bigcirc$\\
\hline 
\end{tabular}\\
$\times$: Not considered,
$\triangle$: Weak,
$\bigcirc$: Good,
$\circledcirc$: Excellent
\end{table}

\begin{table}[t]
\caption{Summary of properties of feature quality measures.}
\label{tab:f_measure}
\centering
\begin{tabular}{l|cccc}
\hline 
& {PC} & {HSIC} & {MI} & {SMI}
\\
\hline \hline 
Non-linear Dependency & $\times$ & $\bigcirc$ & $\bigcirc$ & $\bigcirc$ \\
Model Selection & not needed & $\times$ & $\bigcirc$ & $\bigcirc$ \\
Computational Efficiency & $\circledcirc$ & $\bigcirc$ &  $\times$ & $\triangle$\\
\hline  
\end{tabular}\\
$\times$: Not considered,
$\triangle$: Weak,
$\bigcirc$: Good,
$\circledcirc$: Excellent
\vspace*{5mm}

\caption{Summary of combinations of optimization strategies and feature quality measures.}
\label{tab:opt_measure}
{\footnotesize
\begin{tabular}{c|lp{3cm}lll}
\hline 
 & {Ranking} & {Forward} & {Backward} & {Exhaustive} & {$\ell_1$} \\ \hline \hline 
{PC} & $\bigcirc$\cite{Hall2000} & $\times$ & $\times$ & $\times$ & $\times$ \\
{HSIC} & $-$ &  $\bigcirc$\cite{Song2007}  & $\bigcirc$\cite{Song2007} & $\times$ & $\triangle$\cite{Masaeli2010a} \\
{MI} & $\bigcirc$\cite{Suzuki2008}  & $\bigcirc$ & $\bigcirc$ & $\times$ & $-$ \\
{SMI} & $\bigcirc$\cite{Suzuki2009}  & $\bigcirc$\cite{Suzuki2009}, \cite{Hachiya2010} & $\bigcirc$\cite{Suzuki2009} & $\times$ & $-$\\ \hline   
\end{tabular}
}\\
$\bigcirc$: Method exists,
$\triangle$: Variation exists,\\
$-$: Method does not exist,
$\times$ Method is unreasonable, impractical
\end{table}

\subsection{Formulation of \proposed{} }
\proposed{} attempts to find an $m$-dimensional sparse weight vector by
solving the following optimization problem:
\begin{equation}
\label{eqn:l1lsmi_formu}
\begin{aligned}
& \underset{\w \in \mathbb{R}^m}{\text{maximize}} & & \widehat{I}_s(\diag(\w)\X, \Y) \\
& \text{subject to} & & \bs{1}^T\w \leq r  \\
& & & \w \geq \bs{0},
\end{aligned}
\end{equation}
where $\widehat{I}_s$ is the LSMI defined in \eqref{eqn:lsmi}, $r>0$ is
the radius of the $\ell_1$-ball, $\bs{1}$ is the $m$-dimensional
vector consisting of only 1's, and ``$\geq$'' in $\bs{w} \geq \bs{0}$ is applied
element-wise. Features are selected according to the non-zero coefficients of
the learned $\widehat{\w}$. Here, since the sign of $\widehat{w}_j$ does not
affect the feature selection process, 
we only consider the positive orthant in $\mathbb{R}^m$. Thus, the
constraint $\w \geq \bs{0}$ is imposed, and $\|\w\|_1$ reduces to $\bs{1}^T\w$. 

\subsection{Advantages of \proposed{}}

Using SMI allows a detection of nonlinear dependency between $X$ and $Y$.
Furthermore, by combining it with the $\ell_1$-regularization feature weighting scheme,
feature interaction is also taken into account since all features are considered simultaneously. 
In general, the use of $\ell_1$-regularization does not necessarily give an ability to deal with redundant features. 
That is, the weights of all redundant features may be all high. 
This drawback of $\ell_1$-regularization is covered by the use of SMI. 
Since adding a redundant feature to the selected subset does not increase the SMI value (i.e., no new information), \proposed{} implicitly deals with the feature redundancy issue by avoiding the inclusion of
redundant features. 
This is achieved by simply maximizing SMI between the  weighted
features and the output.
The use of density-ratio estimation in approximating
SMI also helps avoid the density estimation problem, which is
difficult when $m$ is large.

\subsection{Solving \proposed{}}
Here, we explain how we solve the \proposed{} optimization problem.

\subsubsection{Algorithm Overview}
\algoref{algo:l1lsmi_ksearch} is executed to find a
$k$-feature subset by a binary-search-liked scheme. Based on the observation
that the number of obtained features tends to increase as $r$ increases, the
idea is to systematically vary $r$ so that $k$ features can be obtained.
Starting from a low $r$, the \proposed{} optimization problem is solved by
iteratively performing gradient ascent and projection (constraint satisfaction).
If $k$ features can be obtained from the current $r$, then return them.
Otherwise, $r$ is doubled (starting from \algoline{algoline:l1lsmi_ksearch_zh}
in \algoref{algo:l1lsmi_ksearch}) until more than $k$
features are obtained. The value of $r$ firstly found to give more than $k$
features is denoted by $r_{\mathrm{h}}$, and is assumed to be the upper bound of the
value of $r$ which can give $k$ features.
The lower bound $r_{\mathrm{l}}$ is then set to $r_{\mathrm{h}}/2$ which gives strictly less than $k$
features. The rest of the procedure (starting from
\algoline{algoline:l1lsmi_ksearch_bin} in \algoref{algo:l1lsmi_ksearch}) is to find $r \in
(r_{\mathrm{l}},r_{\mathrm{h}})$ using a binary search scheme, so that $k$ features can be obtained. 
In each step of the search, \eqref{eqn:l1lsmi_formu} is solved
using the middle point $r_{\mathrm{m}}$ between $r_{\mathrm{h}}$ and $r_{\mathrm{l}}$. 
If $k$ features cannot be found, $r_{\mathrm{h}}$ or $r_{\mathrm{l}}$ is updated accordingly. This halving procedure
is repeated until $k$ features are found, or the time limit is reached. 

\begin{algorithm}[t]
\caption{Pseudo code of \proposed{} to search for a $k$-feature subset.}
\label{algo:l1lsmi_ksearch}
\begin{algorithmic}[1]
\REQUIRE $k$ (desired number of features)
\STATE $r \leftarrow 0.1$ \COMMENT{$r$ is initially low} 
\REPEAT[try to find an upper bound $r_{\mathrm{h}}$] \label{algoline:l1lsmi_ksearch_zh} 
	\STATE $r \leftarrow 2r$ 
	\STATE $\w_0 \leftarrow $ randomly initialize a feasible $\w$
	\STATE $\mathcal{X}_r \leftarrow$ Solve \eqref{eqn:l1lsmi_formu} with $(\w_0,
	r)$ 
        \COMMENT{$\mathcal{X}_r$: set of features obtained using $r$}
	\IF{$|\mathcal{X}_r| = k$}
		\RETURN $\mathcal{X}_r$
	\ENDIF
\UNTIL{$|\mathcal{X}_r| > k$ or time limit exceeded  }
\STATE $r_{\mathrm{h}} \leftarrow r$
\STATE $r_{\mathrm{l}} \leftarrow r_{\mathrm{h}}/2$
\WHILE[find $r \in (r_{\mathrm{l}}, r_{\mathrm{h}})$ which gives $k$ features with a binary
search]{time limit not exceeded} \label{algoline:l1lsmi_ksearch_bin} 
	\STATE $r_{\mathrm{m}} \leftarrow (r_{\mathrm{h}} + r_{\mathrm{l}})/2$ 
	\STATE $\w_0 \leftarrow $ randomly initialize a feasible $\w$
	\STATE $\mathcal{X}_{r_{\mathrm{m}}} \leftarrow$ Solve \eqref{eqn:l1lsmi_formu} with
	$(\w_0, r_{\mathrm{m}})$ 
	\IF{$|\mathcal{X}_{r_{\mathrm{m}}}| = k$}
		\RETURN $\mathcal{X}_{r_{\mathrm{m}}}$
	\ELSIF{$|\mathcal{X}_{r_{\mathrm{m}}}| < k$}
		\STATE $r_{\mathrm{l}} \leftarrow r_{\mathrm{m}}$
	\ELSIF{$|\mathcal{X}_{r_{\mathrm{m}}}| > k$}
		\STATE $r_{\mathrm{h}} \leftarrow r_{\mathrm{m}}$
	\ENDIF
\ENDWHILE
\STATE $\mathbb{S} \leftarrow $ list of all $\mathcal{X}$ found so far, sorted
in the ascending order by $||\mathcal{X}|-k|, |\mathcal{X}|-k, -\widehat{I}_s(\X_{\mathcal{X}}, \Y)$ \label{algoline:l1lsmi_ksearch_order}
\RETURN the first $\mathcal{X}$ in $\mathbb{S}$ 
\end{algorithmic}

\end{algorithm}

In case that a $k$-feature subset cannot be found, obtained feature subsets
$\mathcal{X}$ are sorted in ascending order of three keys given by
$||\mathcal{X}|-k|, |\mathcal{X}|-k, -\widehat{I}_s(\X_{\mathcal{X}}, \Y)$.
This means that the feature subsets whose size is closest to $k$ are to be put
towards the head of the list. With two sets whose size is
equally closest to $k$, then prefer the smaller one (due to $|\mathcal{X}|-k$). If
there are still many such subsets, bring the ones with highest
$\widehat{I}_s(\X_{\mathcal{X}}, \Y)$ to the head of the list, where
$\X_{\mathcal{X}}$ denotes the data matrix $\X$ with only rows indexed by
$\mathcal{X}$. In the end, the feature subset $\mathcal{X}$ at the head of the
list is selected. 

\subsubsection{Basis Function Design}
Estimation of SMI requires $b$ basis functions.
Here, we choose the basis functions to be in the form of a product kernel defined as
\begin{equation}
\varphi_l(\diag(\w)\x, \y) = \phi_l^x(\diag(\w)\x)\phi_l^y(\y)
\mbox{ for }l=1,\ldots,b. 
\label{eqn:basis_product}
\end{equation}
$\phi_l^x(\cdot)$ is defined to be the Gaussian
kernel,
\begin{equation*}
\phi_l^x(\diag(\w)\x) = \exp\left(-\frac{\|\diag(\w)(\x -
\x_{c(l)})\|^2}{2\sigma^2}\right).
\end{equation*}  
$c(l) \in \{1,\ldots,n\}$ is a randomly chosen sample
index without overlap. The definition of $\phi_l^y(\y)$ depends on the task. For a
regression task, $\phi_l^y(y)$ is also defined to
be a Gaussian kernel,
\begin{equation*}
\phi_l^y(y) = \exp\left(-\frac{ (y -y_{c(l)})^2}{2\sigma^2}\right).
\end{equation*}
For a $C$-class classification task in which $Y \in \{1,\ldots,C\}$, the delta
kernel is used on $\Y$, i.e.,
$\phi_l^y(y)$ takes 1 if $y =
y_{c(l)}$, and 0 otherwise. Using these definitions, model selection
for $(\bs{\varphi}, \lambda)$ is reduced to selecting $(\sigma, \lambda)$.

\subsubsection{Optimization}
Given an initial point $\w_0$ and the radius $r$,
the \proposed{} optimization problem is simply solved by
gradient ascent.
To guarantee the feasibility, the updated $\w$
is projected onto the positive orthant of the constrained $\ell_1$-ball in each iteration. 
The projection can be carried out by first projecting $\w$
onto the positive orthant with
$\max(\w, \bs{0})$,
where the $\max$ function is applied element-wise. This is then followed by a
projection onto the $\ell_1$-ball which can be carried out in $O(m)$ time \cite{Duchi2008}. 

In practice, there are many more sophisticated
methods for solving \eqref{eqn:l1lsmi_formu}, e.g., projected Newton-type methods \cite{Lee2006,Schmidt2007}. These methods generally converge
super-linearly, and are faster (in terms of the convergence rate) than ordinary
gradient ascent algorithms which converge linearly. 
However, the notion of convergence does not take into account the number of
function evaluations. In general, methods with a good convergence rate rely on a
large number of function evaluations per iteration, i.e., performing line search
to find a good step size. In our case, function evaluation is expensive
since model selection for $(\sigma, \lambda)$ has to be performed. It turns out
that using a more sophisticated solver may take more time to actually solve the
problem even though the convergence rate is better. So, we decided to simply use
a gradient ascent algorithm to solve the problem. Additionally, to further improve the
computational efficiency, model selection is performed every five iterations, instead of
every iteration. This is based on the fact that, in each iteration, $\w$ is not
significantly altered. Hence, it makes sense to assume that the selected
$(\sigma, \lambda)$ from the previous iteration are approximately correct.

\section{Experiments}
\label{sec:exp}
In this section, we report experimental results.

\subsection{Methods to be Compared}
We compare the performance of the following feature selection algorithms: 
\begin{itemize}
	\item PC (Pearson correlation ranking).
	\item F-HSIC (forward search with HSIC).
	\item F-LSMI (forward search with LSMI) \cite{Hachiya2010}.
	\item B-HSIC (backward search with HSIC) \cite{Song2007}.
	\item B-LSMI (backward search with LSMI).
	\item $\ell_1$-HSIC (similar to \proposed{}, but the objective function is replaced
	with $\text{HSIC}(\diag(\w)\X,\Y)$) .
	\item \proposed{}\footnote{Matlab implementation of \proposed{} is available at \url{http://wittawat.com/software/l1lsmi/}} (proposed method).
	\item mRMR (Minimum Redundancy Maximum Relevance) \cite{Peng2005}. 
	mRMR is one of the state-of-the-art algorithms which selects features by solving 
\begin{equation*}
\begin{aligned}
& \underset{\mathcal{I} \subset \{1,\ldots,m\}}{\text{maximize}}
& & \overbrace{\frac{1}{k} \sum_{i \in \mathcal{I}} I(X_i, Y)}^{\text{relevancy measure}}  
 - \overbrace{\frac{1}{k^2} \sum_{i \in
\mathcal{I}} \sum_{j \in \mathcal{I}} I(X_i, X_j)}^{\text{redundancy measure}} \\
& \text{subject to}
& & |\mathcal{I}| = k.
\end{aligned}
\end{equation*}
	That is, it uses mutual information to select relevant features which are not too redundant.
	mRMR solves the optimization problem by greedily adding one feature at a time until $k$ features
	can be obtained. This scheme is similar to a forward search algorithm.
	
	\item QPFS (Quadratic Programming Feature Selection) \cite{Rodriguez-Lujan2010}.
	QPFS formulates the feature selection task as a quadratic programming problem of the form:
	\begin{equation*}
	\begin{aligned}
	& \underset{\w \in \mathbb{R}^m}{\text{minimize}} & & 
	 \frac{1}{2}(1-\alpha)\w^T \bs{Q}\w - \alpha \bs{f}^T \w\\
	& \text{subject to} & & \bs{1}^T\w = 1  \\
	& & & \w \geq \bs{0},
	\end{aligned}
	\end{equation*} 	
	where $0 \leq \alpha \leq 1$ controls the trade-off between high relevancy (high $\alpha$) and low redundancy 
	of the selected features. $\bs{Q} = [q_{ij}] = |\rho(X_i, X_j)|$ is the absolute value of the
	Pearson correlation between $X_i$ and $X_j$ as in \eqref{eqn:pearson}, and $\bs{f} = [f_i] = |\rho(X_i, Y)|$.
	In the case that $Y$ is categorical, the correlation for categorical variable as in \cite{Hall2000} is used.
	In this experiment, we use the recommended value of $\alpha = \bar{q}/(\bar{q}+\bar{f})$ where 
	$\bar{q} = \frac{1}{m^2}\sum_{i=1}^m \sum_{j=1}^m q_{ij}$ and $\bar{f} = \frac{1}{m}\sum_{i=1}^m f_i$ \cite{Rodriguez-Lujan2010}.
	Notice that if $\alpha=1$, QPFS reduces to PC.	
	
	\item Lasso \cite{Tibshirani1996}. Lasso is a well-known method of least squares which imposes an $\ell_1$-norm
          constraint on the weight vector. Specifically, it solves the problem of the form:
	\begin{equation*}
	\begin{aligned}
	& \underset{\w \in \mathbb{R}^m}{\text{minimize}} & & 
	 \|\Y - \w^T \X\|^2 + \lambda \|\w\|_1, \\
	\end{aligned}
	\end{equation*} 	
	where $\lambda \geq 0 $ is the sparseness regularization parameter. In this experiment, $\lambda$ is varied
	so that $k$ features can be obtained. 
	
	\item Relief \cite{Kira1992,Kononenko1994}.
	Relief is another state-of-the-art heuristic algorithm which scores each feature based 
	on how it can discriminate different classes (distance-based).
\end{itemize}

\subsection{Toy Data Experiment}
An experiment is conducted on the following three toy datasets:
\begin{enumerate}
\item \texttt{and-or}
	\begin{itemize}
		\item Binary classification (4 true / 6 distracting features).
		\item $Y = (X_1 \wedge X_2) \vee (X_3 \wedge X_4) $.
		\item $X_1, \ldots, X_7 \sim \text{Bernoulli(0.5)}$,
                  where $\mathrm{Bernoulli}(p)$ denotes the Bernoulli distribution
                  taking value $1$ with probability $p$.
		\item $X_8,\ldots, X_{10} = Y$ with 0.2 chance of bit flip.
		\item {Characteristics:} Feature redundancy and weak interaction.    
	\end{itemize}

\item \texttt{quad}
	\begin{itemize}
	\item Regression (2 true / 8 distracting features).
	\item $Y = \frac{X_1^2 + X_2}{0.5 + (X_2 + 1.5)^2} + 0.1\epsilon$.
	\item $X_1,\ldots, X_8,\epsilon \sim \mathcal{N}(0,1)$,
          where $\mathcal{N}(\mu,\sigma^2)$ denotes the normal distribution
          with mean $\mu$ and variance $\sigma^2$.
	\item $X_9 \sim 0.5X_1 + \mathcal{U}(-1, 1)$,
          where $\mathcal{U}(a,b)$ is the uniform distribution on $[a,b]$.
	\item $X_{10} \sim 0.5X_2 + \mathcal{U}(-1, 1)$.
	\item {Characteristic:} Non-linear dependency. 
	\end{itemize}

\item \texttt{xor}
	\begin{itemize}
	\item Binary classification (2 true / 8 distracting features).
	\item $Y = \text{xor}(X_1,X_2)$, where $\text{xor}(X_1,X_2)$
          denotes the XOR function for $X_1$ and $X_2$.
	\item $X_1,\ldots,X_5 \sim \text{Bernoulli(0.5)}$. 
	\item $X_6,\ldots,X_{10} \sim \text{Bernoulli(0.75)}$. 
	\item {Characteristic:} Feature interaction.     
	\end{itemize}
\end{enumerate}	
      
The number of features to select, $k$, is set to the number of true features in the
respective dataset. For LSMI-based methods, Gaussian kernels are used as
the basis functions and $b$ is set to 100. Five-fold cross
validation is carried out on a grid of $(\sigma, \lambda)$
candidates for model selection. For $\sigma$, the candidates are also adaptively
scaled with the median of pairwise sample distance $\sigma_{\mathrm{med}}$, which depends on the currently
selected features.
\begin{equation*}
\sigma_{\mathrm{med}} = \median( \{\|\x_i - \x_j\|_2\}_{i<j}).
\end{equation*}
Gaussian kernels are also used in HSIC-based methods. However, since model selection is not
available for HSIC, in F-HSIC and B-HSIC, the Gaussian width is heuristically set
to $\sigma_{\mathrm{med}}$ \cite{Scholkopf2002}. For $\ell_1$-HSIC, the
Gaussian width is adaptively set to the median of pairwise distance of $\diag(\w)\X$ every
five iterations. Due to the non-convexity of the objective functions,
\proposed{} and $\ell_1$-HSIC are restarted 20 times with 
randomly chosen initial points.

The experiment is repeated 50 times with $n=400$ points sampled in each trial.
For each method and each dataset, an average of the F-measure over all trials is
reported. The F-measure is defined as $f = 2pr/(p+r)$, where
\begin{itemize}
\item $p = $ (number of correctly selected features) / (number of selected
features).
\item $r = $ (number of correctly selected features) / (number of correct
features).
\end{itemize}
An F-measure is bounded between 0 and 1, and 1 is achieved if and
only if all the true features are selected and none of the distracting features
is selected. The results are shown in 
\tabref{tab:fmeasure}.
 
\begin{table*}[tb]
\caption{Averaged F-measures on the \texttt{and-or},
  \texttt{quad}, and \texttt{xor} datasets.}
\label{tab:fmeasure}
{ \hspace*{2mm}
\begin{tabular}{l|lllll}
\hline
Dataset & PC & F-HSIC & F-LSMI & B-HSIC & B-LSMI  \\ \hline \hline 
\texttt{and-or} & 0.25 (.00) & 0.25 (.00) & 0.57 (.22) & 0.25 (.00) & 0.85 (.22)\\ 
\texttt{quad} & 0.57 (.20) & 0.95 (.15) & \textbf{1.00 (.00)} & 0.95 (.15) & \textbf{1.00 (.00)} \\ 
\texttt{xor} & 0.25 (.31) & 0.52 (.50) & 0.53 (.50) & \textbf{1.00 (.00)} & \textbf{1.00 (.00)}\\ \hline 
\end{tabular}
}
\vspace{3mm}
{ \hspace*{13mm} 
\begin{tabular}{l|llllll}
\hline
Dataset &  $\ell_1$-HSIC & $\ell_1$-LSMI & mRMR & QPFS & Lasso & Relief \\ \hline \hline 
\texttt{and-or} &  0.25 (.00) & \textbf{1.00 (.00)} & 0.25 (.00) & 0.41 (.17) & 0.21 (.09) & 0.55 (.15) \\ 
\texttt{quad} & 0.64 (.23) & \textbf{1.00 (.00)} & \textbf{1.00 (.00)} & 0.64 (.23) & 0.66 (.25) & \textbf{1.00 (.00)} \\ 
\texttt{xor} & \textbf{1.00 (.00)} & \textbf{1.00 (.00)} & 0.28 (.31) & 0.25 (.32) & 0.26 (.32) & \textbf{1.00 (.00)} \\ \hline 
\end{tabular}
}

\end{table*}


PC ranks the relevance of each feature individually without taking into account the redundancy among features. 
This results in a failure on the \texttt{and-or} dataset
since $X_8,\ldots,X_{10}$, which are redundant, would simply be ranked
top due to their similarity to $Y$. 

The forward search variants do not work on problems with feature interaction. 
To detect interacting features, it is necessary that all features be considered simultaneously.
For this reason, F-HSIC and F-LSMI fail in the \emph{xor} problem.

The performance of HSIC-based methods seems to be unstable in many cases.
A possible cause of the instability is from the use of an incorrect parameter: 
The heuristic of using $\sigma_{\mathrm{med}}$ for the Gaussian width does not always work. 
As an example, given a fixed data matrix $\X$, the more features selected, the larger $\sigma_{\mathrm{med}}$ may become. 
This is because the Euclidean distance is a non-decreasing function of the dimension. 
So, inclusion of many irrelevant features obviously unnecessarily makes $\sigma_{\mathrm{med}}$ larger. 
B-HSIC is subject to this weakness since it starts the search with all features. 

B-LSMI performs well in detecting non-linear dependency (\texttt{quad})
and feature interaction (\texttt{xor}).
However, due to its greedy nature, the redundant features in the \texttt{and-or} problem
are sometimes chosen.
That is, in the first few iterations, all redundant features are kept, and one of the true features is eliminated instead.

mRMR and QPFS have similar optimization strategies. 
That is, both of them measure the relevancy of each feature, and have a 
pairwise feature redundancy constraint. Regardless of the feature measure in use, 
considering features in a univariate way cannot reveal interacting features (by definition of feature interaction). 
Therefore, it is not surprising that both of them fail on the \texttt{xor} and \texttt{and-or} datasets.
Nevertheless, mRMR works well on the \texttt{quad} dataset since
mutual information can reveal a non-linear dependency. On the other hand, QPFS and Lasso
do not perform well on the \texttt{quad} dataset since both of them use a linear measure. 

Relief is one of the few feature ranking algorithms which can consider feature interaction (the \emph{xor} dataset) because of its distance-based nature. However, it suffers the same drawback
as other ranking algorithms in that no redundancy is considered. Hence, it fails on the \emph{and-or} dataset with the same reason as PC. 

The proposed \proposed{} performs well on all datasets. 
This clearly shows that \proposed{} can consider redundancy, detect non-linear dependency, and consider feature interaction.
$\ell_1$-based feature optimization enables a simultaneous consideration of features, which is the key
in tackling the feature interaction problem. 
By using $\ell_1$-regularization in combination with SMI which can detect a
non-linear dependency, \proposed{} can correctly choose the two true features in
the \texttt{quad} problem. For the \texttt{and-or} problem, the pitfall is to choose
$X_8,\ldots,X_{10}$ because of their high correlation to $Y$. However, due to the
usage of $\ell_1$-regularization, \proposed{} attempts to find the four-feature subset which
maximizes LSMI in a non-greedy manner. Since $X_8,\ldots,X_{10}$ contain
bit-flip noise, inclusion of any of them will not deliver the maximum LSMI. In
this case, the only four features which give the maximum LSMI are ${X_1,\ldots,
X_4}$, and thus preferred over any of $X_8,\ldots,X_{10}$. 

As an illustration of LSMI, \tabref{tab:andor_comb74} shows all possible 35
four-feature subsets of $\{X_1,\ldots,X_4\} \cup \{X_8,\ldots,X_{10}\}$ in
the \texttt{and-or} problem and their corresponding LSMI values. It is evident that the
correct subset $\{X_1,\ldots, X_4\}$ has the highest LSMI.
Inclusion of any of $X_8,\ldots,X_{10}$ (and thus remove some from
$\{X_1,\ldots,X_4\}$) would cause a significant drop of the LSMI value. In the
extreme case, with all $X_8,\ldots,X_{10}$ in the selected set (shown at the
bottom of the table), the LSMI score becomes considerably low. This is because each
of $X_8,\ldots,X_{10}$ contains roughly the same information to explain $Y$. Thus,
there is no gain in adding more features which share very similar information.

\begin{table}[t]
\caption{All possible 35 four-feature subsets of $\{X_1,\ldots,X_4\} \cup
\{X_8,\ldots,X_{10}\}$ in  the \texttt{and-or} dataset, and their corresponding
values of LSMI to the output $Y = (X_1 \wedge X_2) \vee (X_3 \wedge X_4) $.}
\label{tab:andor_comb74}
\centering
\begin{minipage}[b]{0.4\textwidth}
\centering
\begin{tabular}{cccc|c}
 \hline
\multicolumn{4}{c|}{Feature indices} & LSMI \\
  \hline \hline
1 & 2 & 3 & 4 &   \textbf{0.496} \\ 
1 & 2 & 3 & 8 &     0.365 \\ 
1 & 2 & 3 & 9 &     0.381 \\ 
1 & 2 & 3 & 10 &     0.357 \\ 
1 & 2 & 4 & 8 &     0.376 \\ 
1 & 2 & 4 & 9 &     0.384 \\ 
1 & 2 & 4 & 10 &     0.372 \\ 
1 & 2 & 8 & 9 &     0.346 \\ 
1 & 2 & 8 & 10 &     0.330 \\ 
1 & 2 & 9 & 10 &     0.336 \\ 
1 & 3 & 4 & 8 &     0.382 \\ 
1 & 3 & 4 & 9 &     0.376 \\ 
1 & 3 & 4 & 10 &     0.392 \\ 
1 & 3 & 8 & 9 &     0.325 \\ 
1 & 3 & 8 & 10 &     0.330 \\ 
1 & 3 & 9 & 10 &     0.333 \\
1 & 4 & 8 & 9 &     0.342 \\ \hline
\end{tabular}
\end{minipage}
~~
\begin{minipage}[b]{0.4\textwidth}
\centering
\begin{tabular}{cccc|c}
 \hline
\multicolumn{4}{c|}{Feature indices} & LSMI \\
  \hline \hline
1 & 4 & 9 & 10 &     0.341 \\ 
2 & 3 & 4 & 8 &     0.367 \\ 
2 & 3 & 4 & 9 &     0.382 \\ 
2 & 3 & 4 & 10 &     0.390 \\ 
2 & 3 & 8 & 9 &     0.341 \\ 
2 & 3 & 8 & 10 &     0.312 \\ 
2 & 3 & 9 & 10 &     0.322 \\ 
2 & 4 & 8 & 9 &     0.340 \\ 
2 & 4 & 8 & 10 &     0.328 \\ 
2 & 4 & 9 & 10 &     0.328 \\ 
3 & 4 & 8 & 9 &     0.356 \\ 
3 & 4 & 8 & 10 &     0.349 \\ 
3 & 4 & 9 & 10 &     0.353 \\  \hline
1 & 8 & 9 & 10 &     0.330 \\
2 & 8 & 9 & 10 &     0.334 \\   
3 & 8 & 9 & 10 &     0.303 \\ 
4 & 8 & 9 & 10 &     0.335 \\ \hline 
\end{tabular}
\end{minipage}
\end{table}

\subsection{Real-Data Experiment}
To demonstrate the practical use of the proposed \proposed{}, we conduct
experiments on real datasets without any specific domains.
All the real datasets used in the experiments are summarized in \tabref{tab:real_dataset}.
The ``Task'' column denotes the type of the problem (R for regression, and C$x$
for $x$-class classification problem). The datasets cover a wide range of
domains including image, speech, and bioinformatics.

\begin{table}[t]
\caption{Summary of the real datasets used in the experiments.}
\label{tab:real_dataset}
\centering
\begin{tabular}{l|rrl|l}
\hline 
Dataset & \multicolumn{1}{l}{~~~~$m$} & \multicolumn{1}{l}{~~~~~~~$n$} & Task &
Class balance (\%) \\ 
  \hline \hline
abalone & 8 & 4177 & R & -  \\ 
bcancer & 9 & 277 & C2 & 70.8/29.2   \\ 
cpuact & 21 & 3000 & R & -  \\
ctslices & 379 & 53500 & R & - \\  
flaresolar & 9 & 1066 & C2 & 44.7/55.3  \\ 
german & 20 & 1000 & C2 & 70.0/30.0  \\ 
glass & 9 & 214 & C6 & 32.7/35.5/7.9/6.1/4.2/13.6 \\  
housing & 13 & 506 & R & -  \\ 
image & 18 & 1155 & C2 & 42.9/57.1  \\ 
ionosphere & 33 & 351 & C2 & 64.1/35.9  \\  
isolet & 617 & 6238 & C26 & about 3.85\% per class \\  
msd & 90 & 10000 & R & - \\ 
musk1 & 166 & 476 & C2 & 56.5/43.5  \\ 
musk2 & 166 & 6598 & C2 & 84.6/15.4  \\ 
satimage & 36 & 6435 & C6 & 23.8/10.9/21.1/9.7/11.0/23.4  \\ 
segment & 18 & 2310 & C7 & 14.3\% per class  \\ 
senseval2 & 50 & 534 & C3 & 33.3\% per class \\ 
sonar & 60 & 208 & C2 & 46.6/53.4  \\ 
spectf & 44 & 267 & C2 & 20.6/79.4  \\ 
speech & 50 & 400 & C2 & 50.0/50.0 \\
vehicle & 18 & 846 & C4 & 25.1/25.7/25.8/23.5  \\ 
vowel & 13 & 990 & C11 & 9.1\% per class  \\ 
wine & 13 & 178 & C3 & 33.1/39.9/27.0  \\ \hline 
\end{tabular}\\
All datasets were taken from UCI Machine Learning Repository:
\url{http://archive.ics.uci.edu/ml/},
except that \texttt{cpuact} is from
\url{http://mldata.org/repository/data/viewslug/uci-20070111-cpu_act/},
\texttt{SENSEVAL-2} is from the Second International Workshop on
Evaluating Word Sense Disambiguation Systems: 
\url{http://www.sle.sharp.co.uk/senseval2},
and \texttt{speech} is our In-house developed voice dataset.
\end{table}

\begin{figure*}[tbp]
\centering

\subfloat[image]{
	\includegraphics[width=0.32\textwidth]{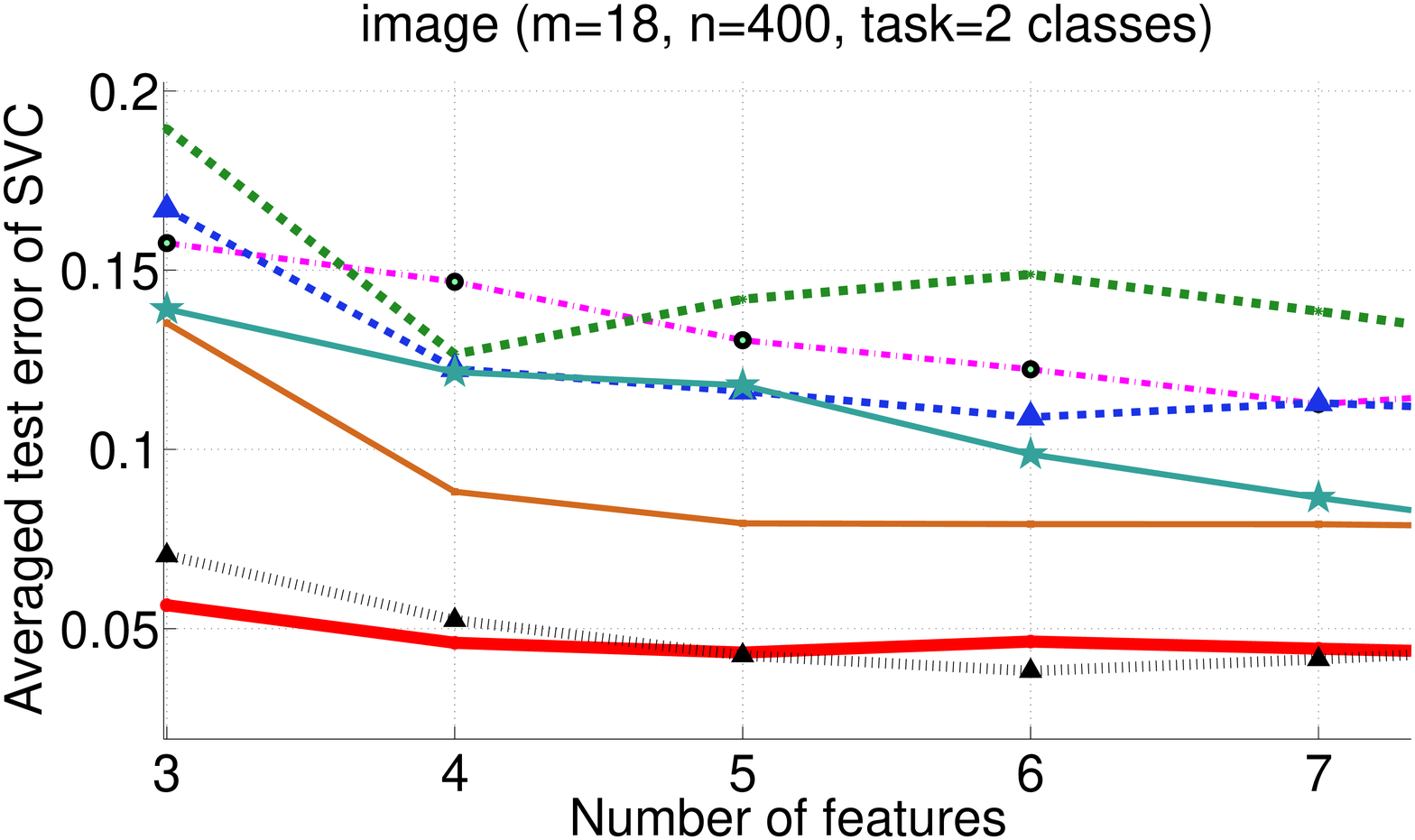}
	\label{fig:fvserr_image_noseq}
}
\subfloat[german]{
	\includegraphics[width=0.32\textwidth]{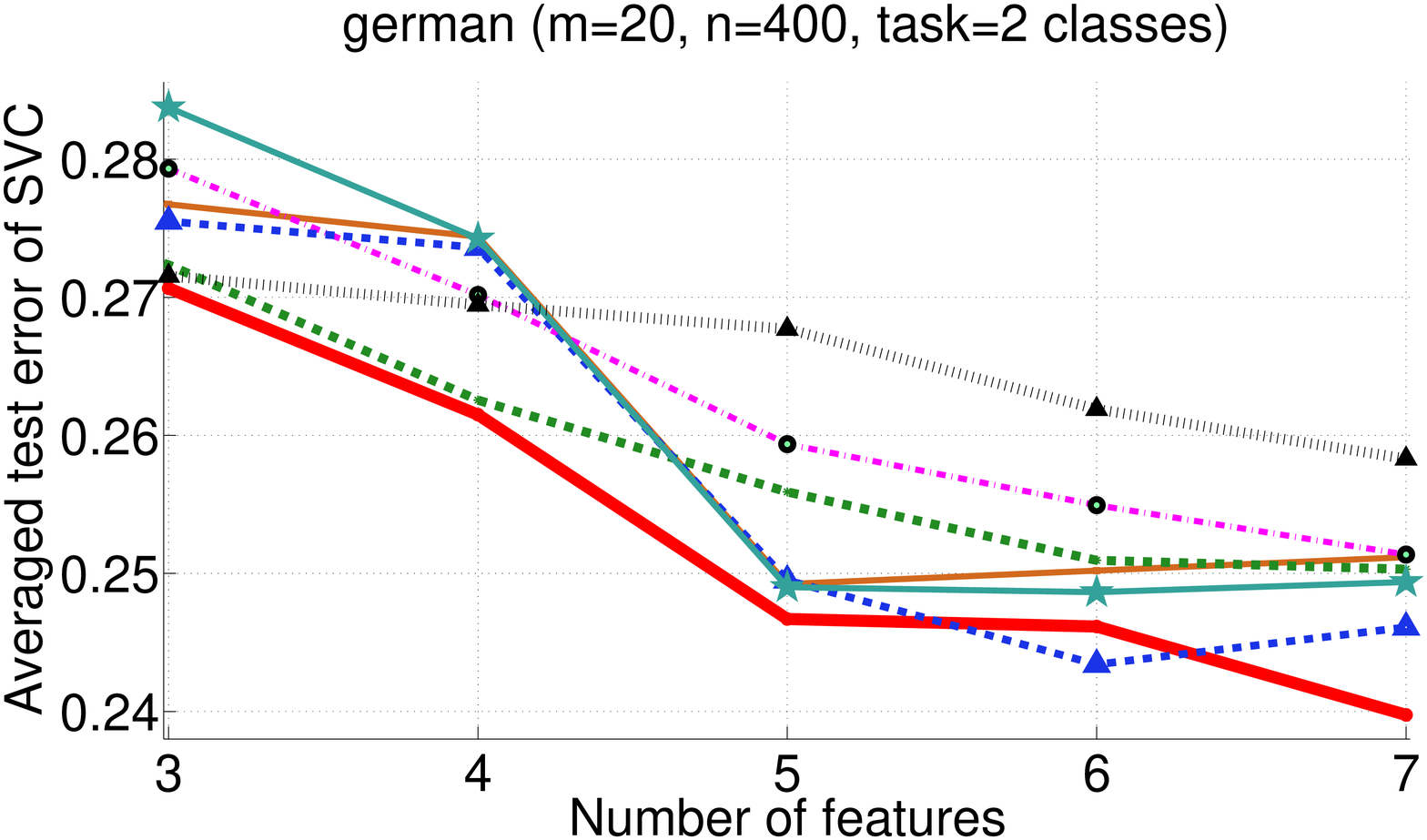}
	\label{fig:fvserr_german_noseq}
}
\subfloat[cpuact]{
	\includegraphics[width=0.32\textwidth]{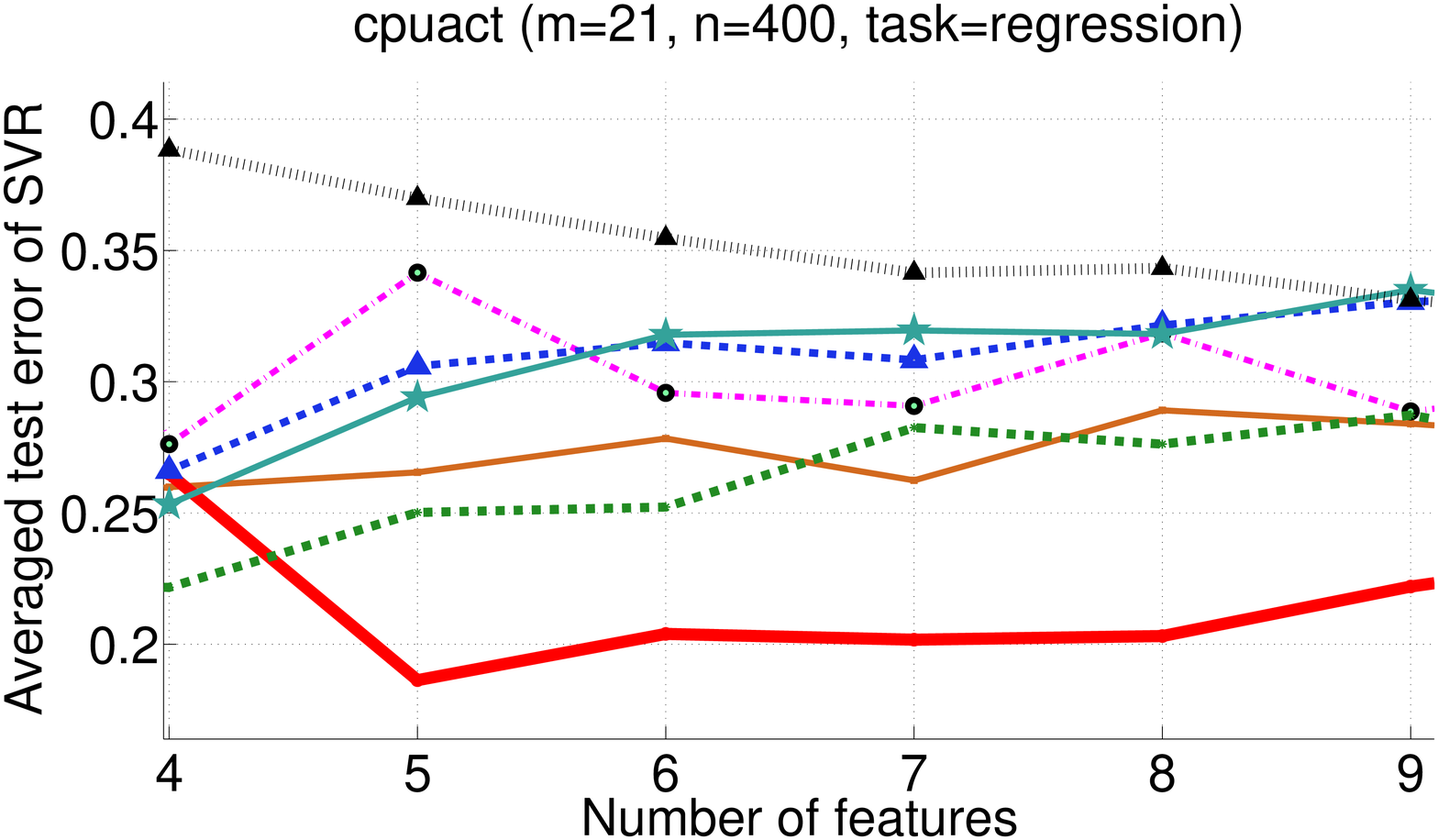}
	\label{fig:fvserr_cpuact_noseq}
}\\
\subfloat[segment]{
	\includegraphics[width=0.32\textwidth]{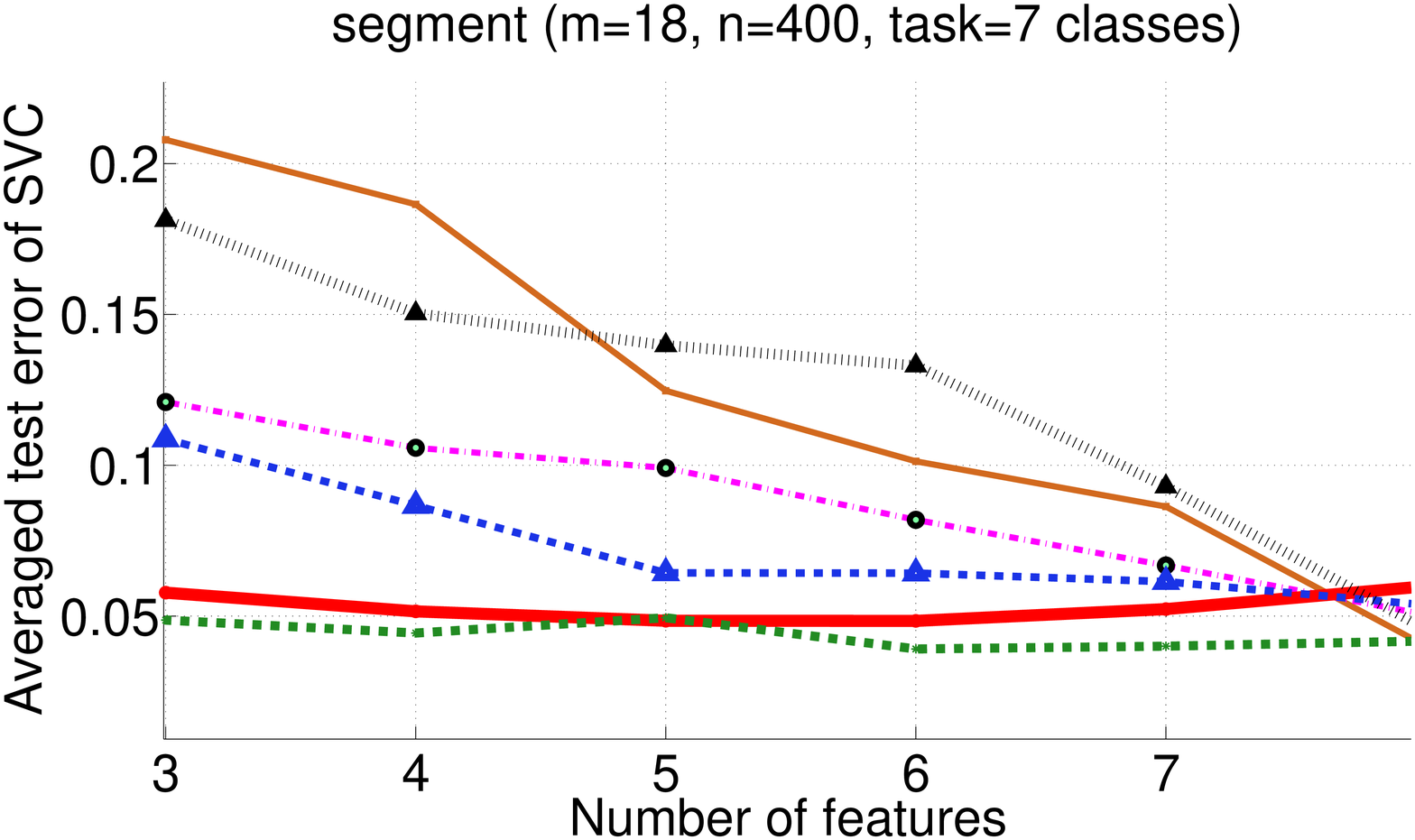}
	\label{fig:fvserr_segment_noseq}
}
\subfloat[wine]{
	\includegraphics[width=0.32\textwidth]{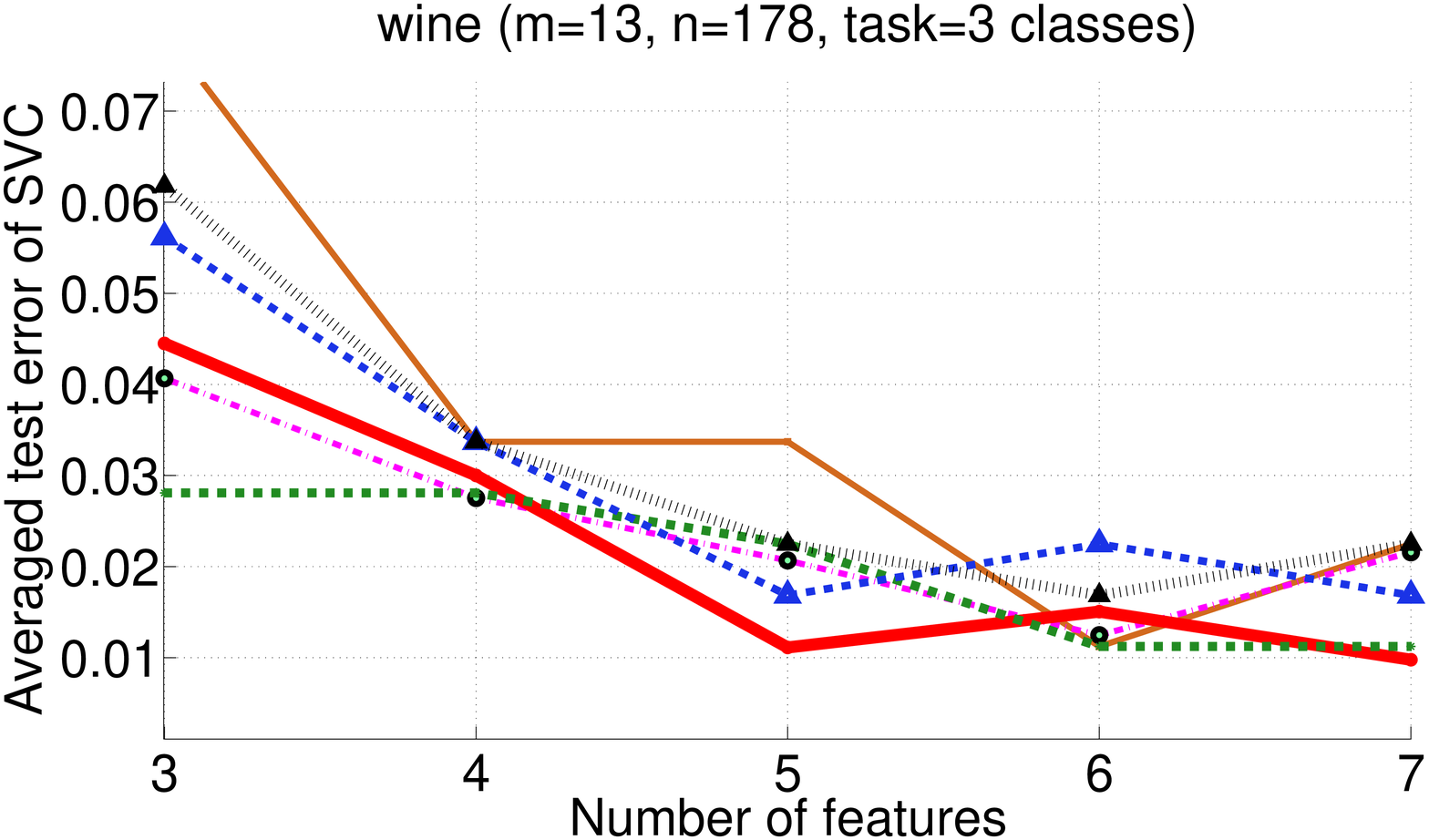}
	\label{fig:fvserr_wine_noseq}
}
\subfloat[flaresolar]{
	\includegraphics[width=0.32\textwidth]{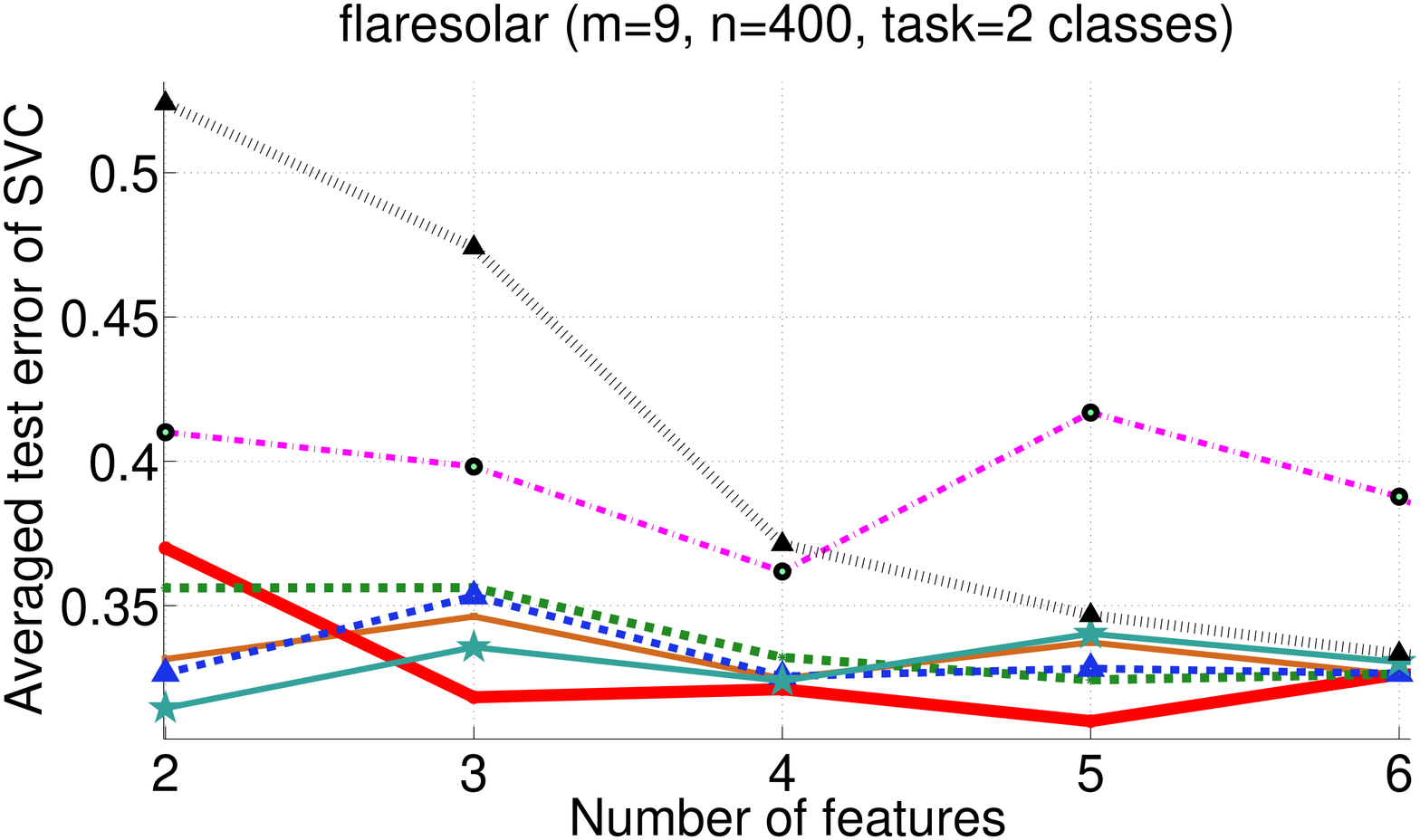}
	\label{fig:fvserr_flaresolar_noseq}
}\\

\subfloat[spectf]{
	\includegraphics[width=0.32\textwidth]{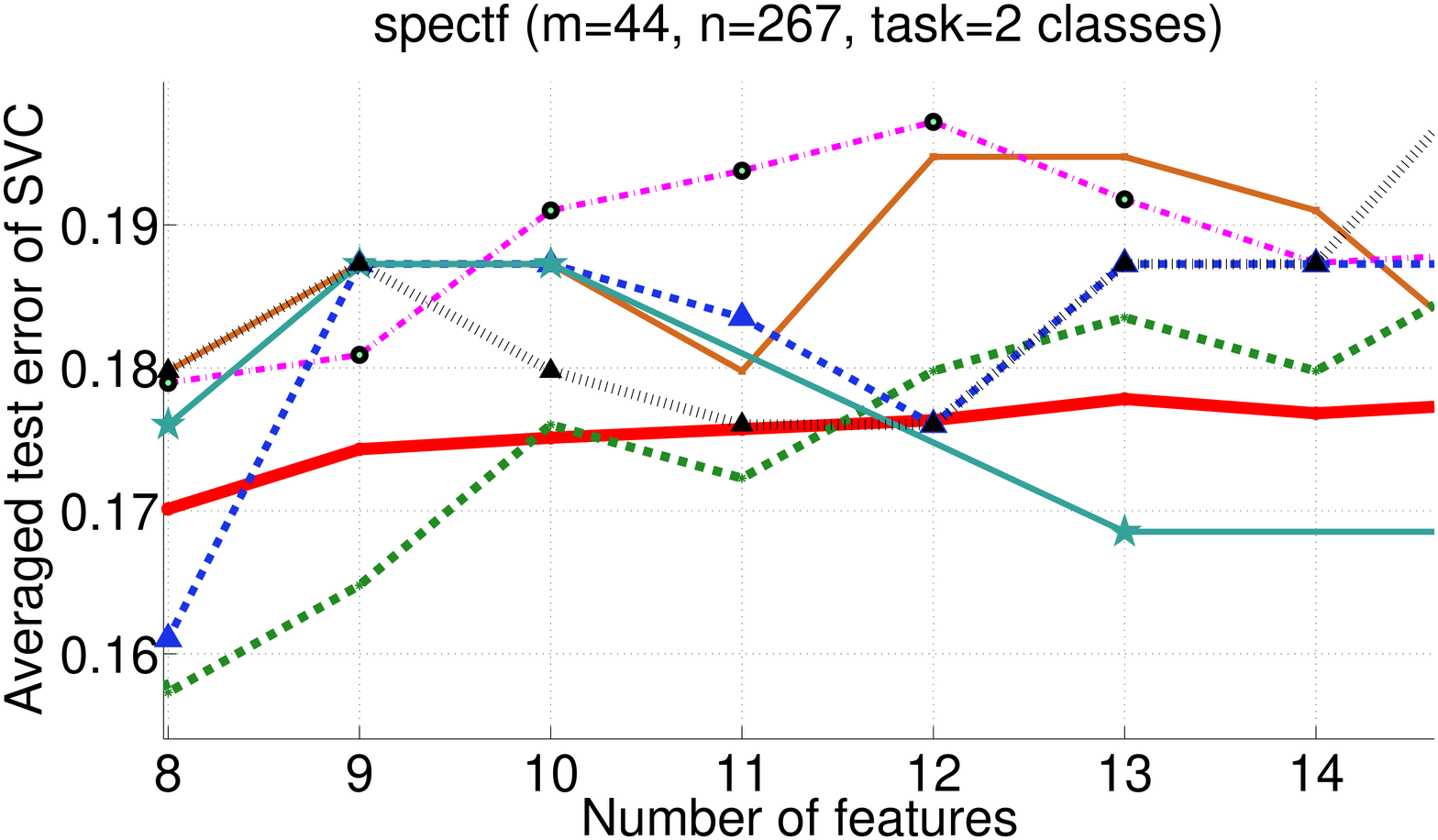}
	\label{fig:fvserr_spectf_noseq}
}
\subfloat[satimage]{
	\includegraphics[width=0.32\textwidth]{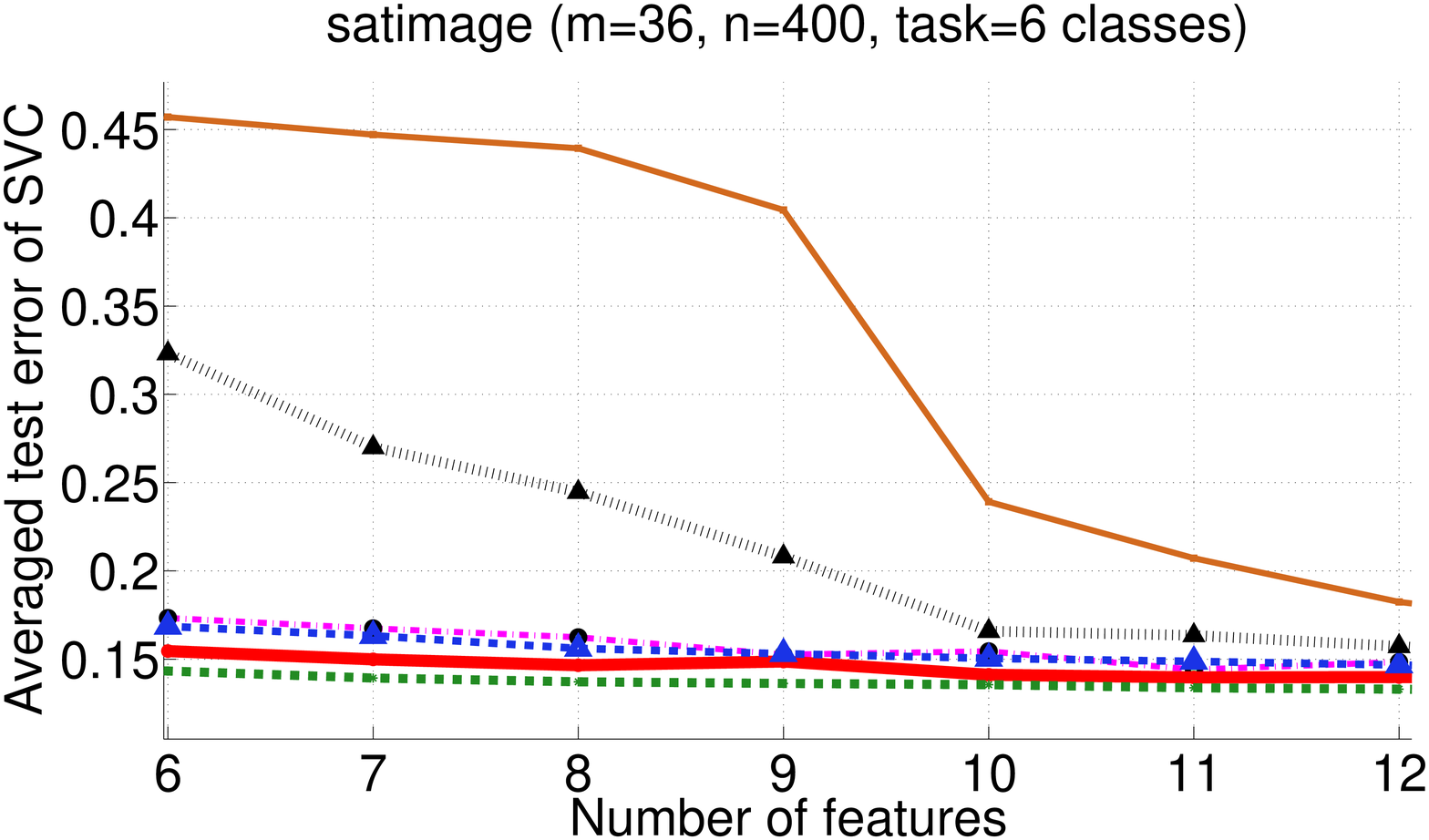}
	\label{fig:fvserr_satimage_noseq}
}
\subfloat[vehicle]{
	\includegraphics[width=0.32\textwidth]{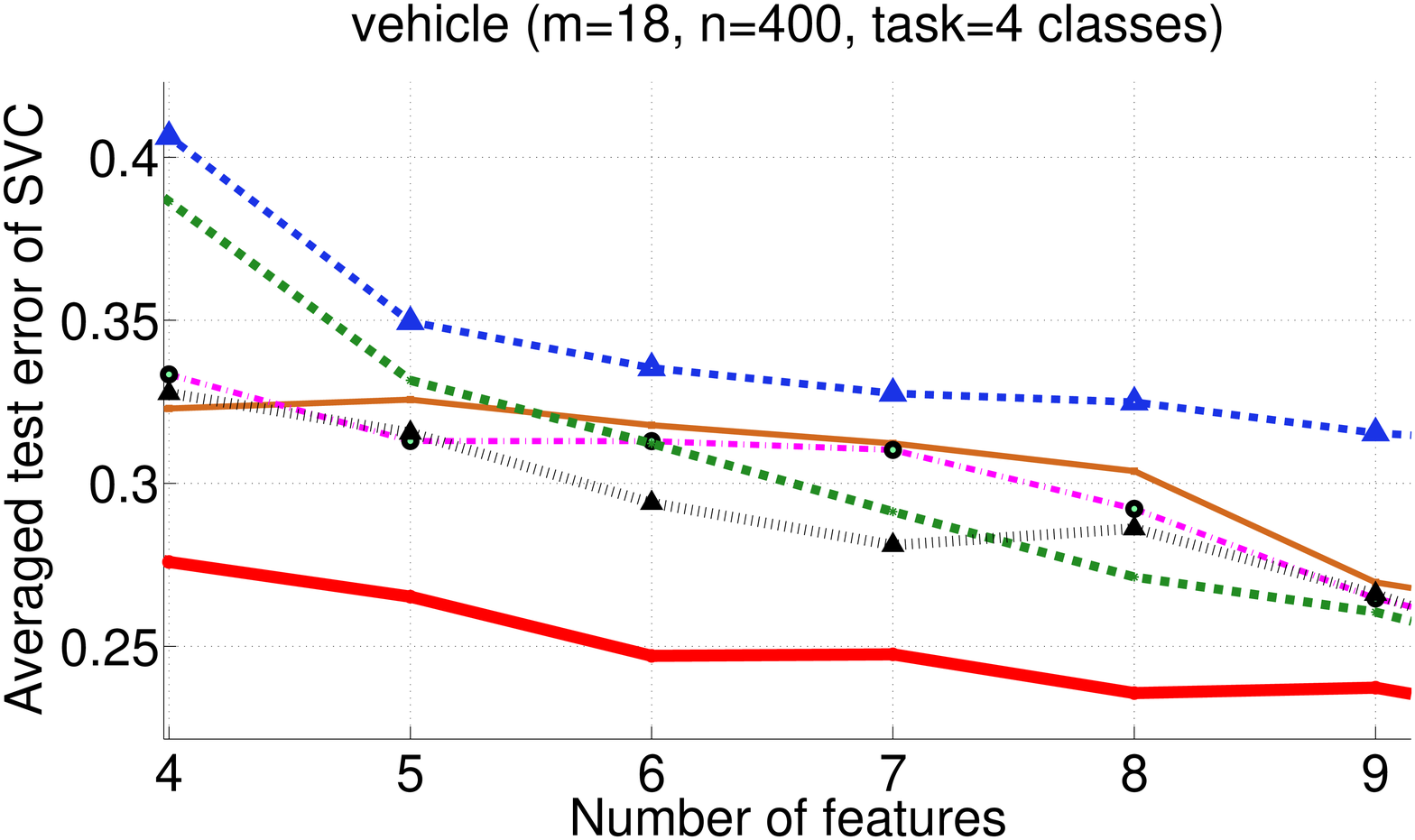}
	\label{fig:fvserr_vehicle_noseq}
}\\
\subfloat[sonar]{
	\includegraphics[width=0.32\textwidth]{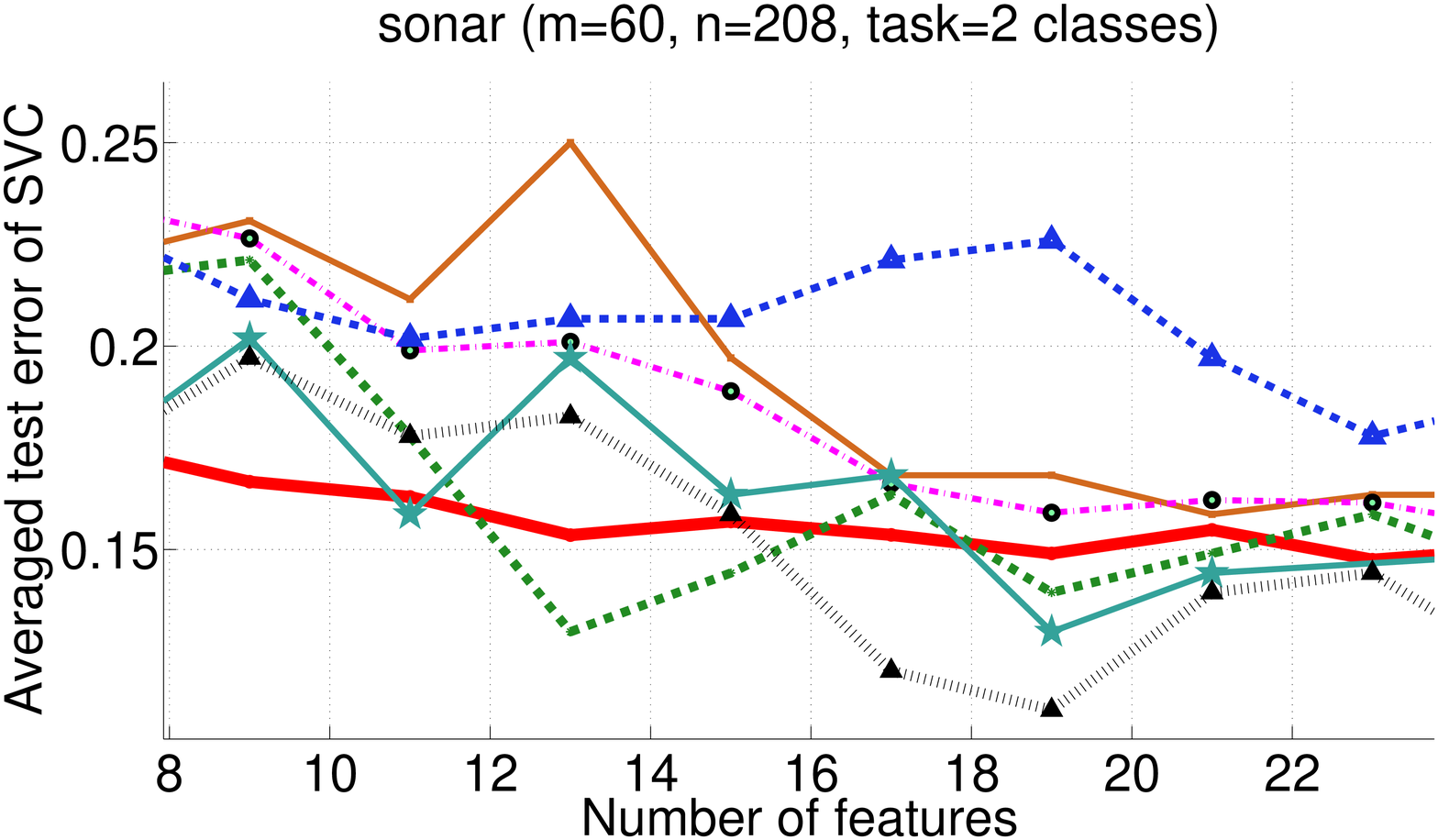}
	\label{fig:fvserr_sonar_noseq}
}
\subfloat[speech]{
	\includegraphics[width=0.32\textwidth]{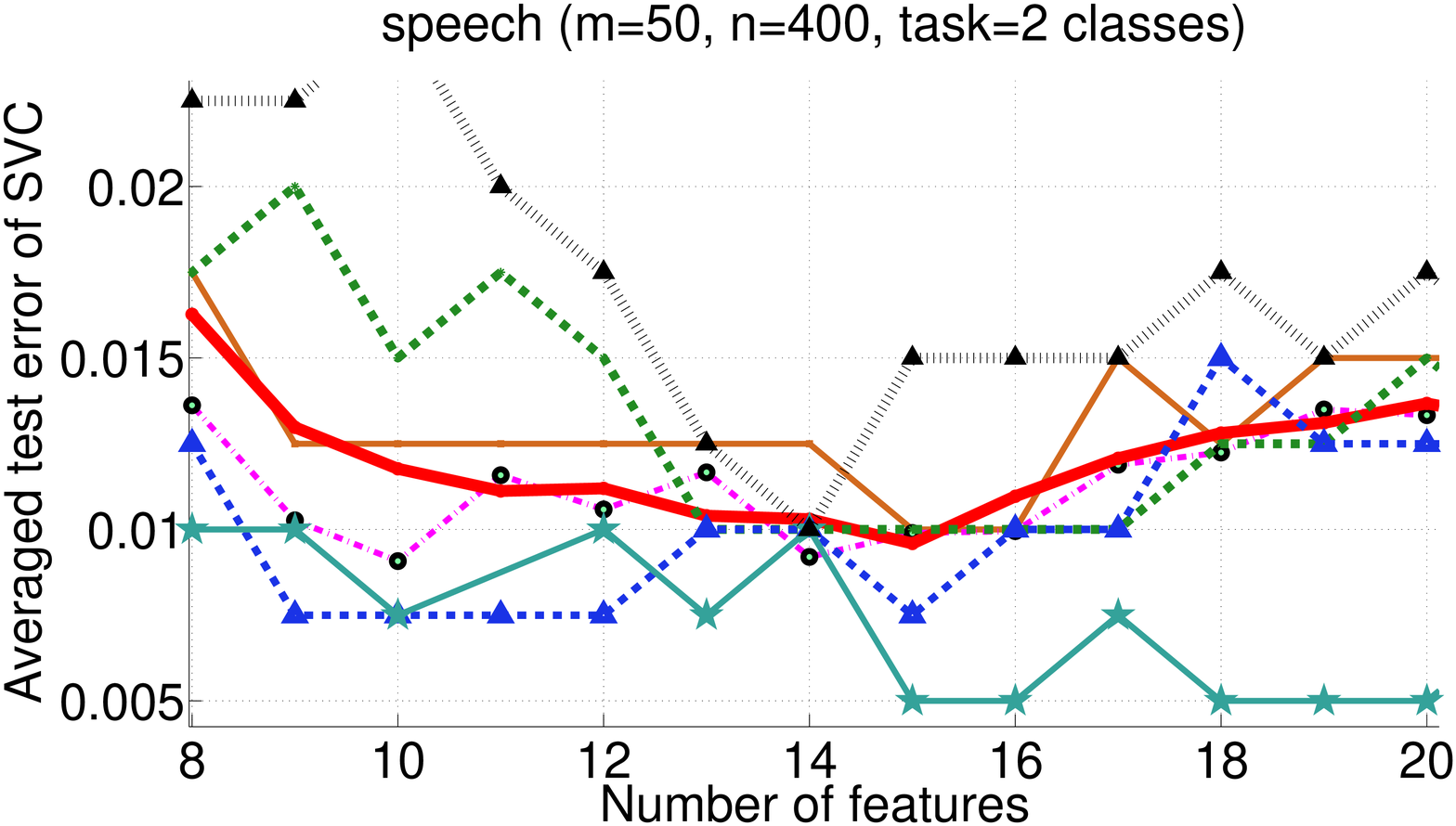}
	\label{fig:fvserr_speech_noseq}
}
\subfloat[senseval2]{
	\includegraphics[width=0.32\textwidth]{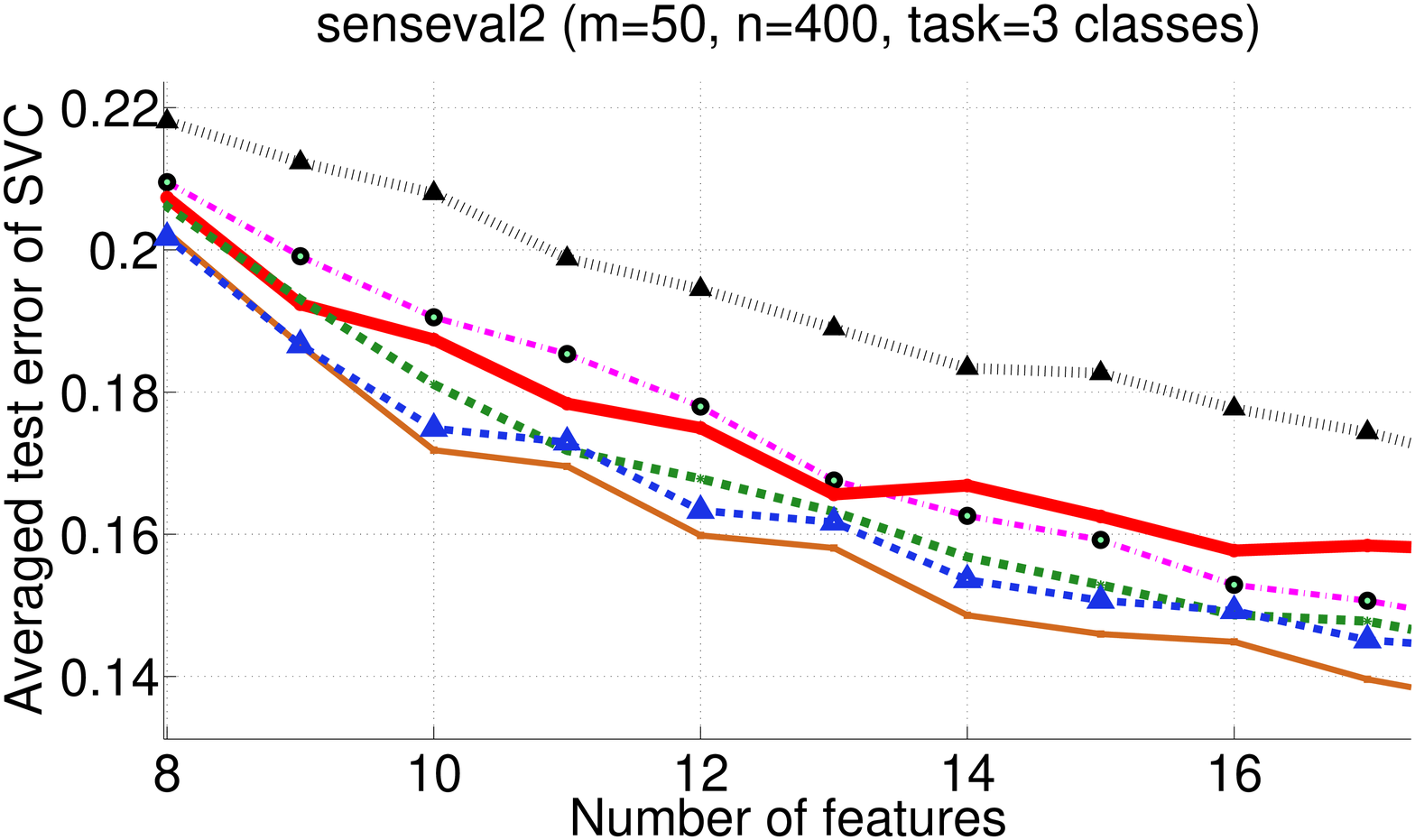}
	\label{fig:fvserr_senseval2_noseq}
}\\
\subfloat[musk1]{
	\includegraphics[width=0.32\textwidth]{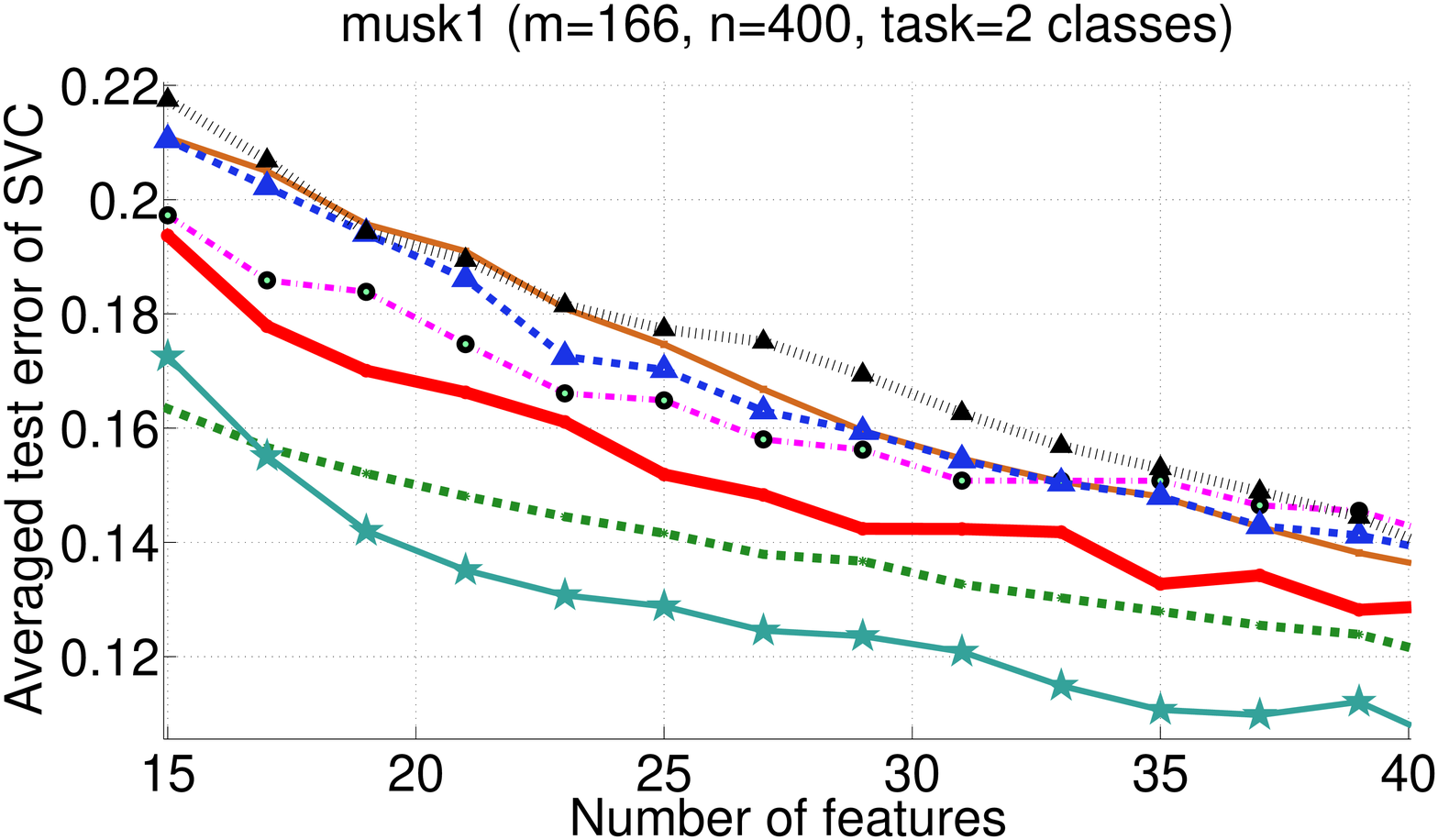}
	\label{fig:fvserr_musk1_noseq}
}
\subfloat[musk2]{
	\includegraphics[width=0.32\textwidth]{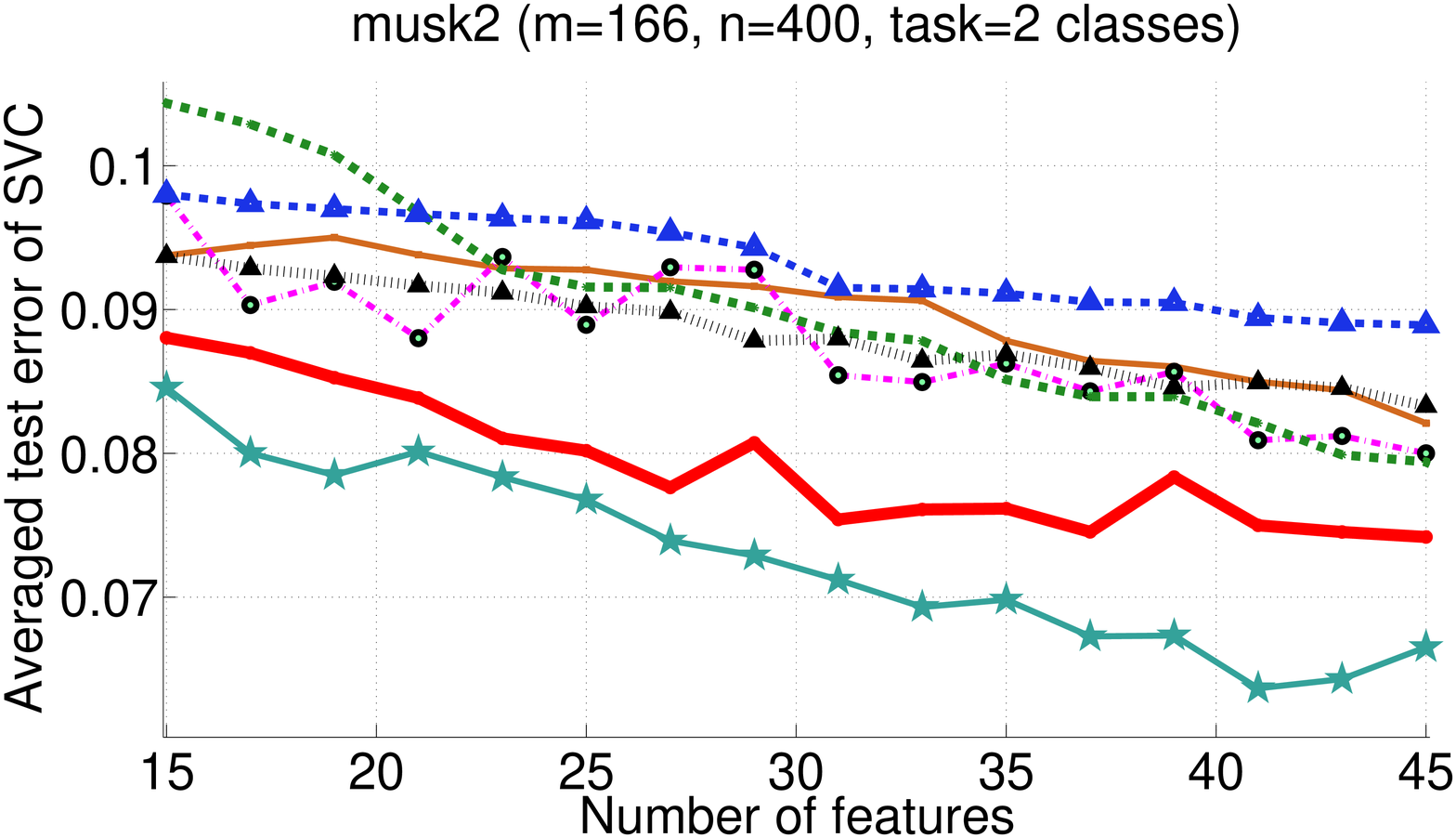}
	\label{fig:fvserr_musk2_noseq}
}
\subfloat{
~~~~~~~~~~~~~~~~~~~~~
\includegraphics[width=0.14\textwidth]{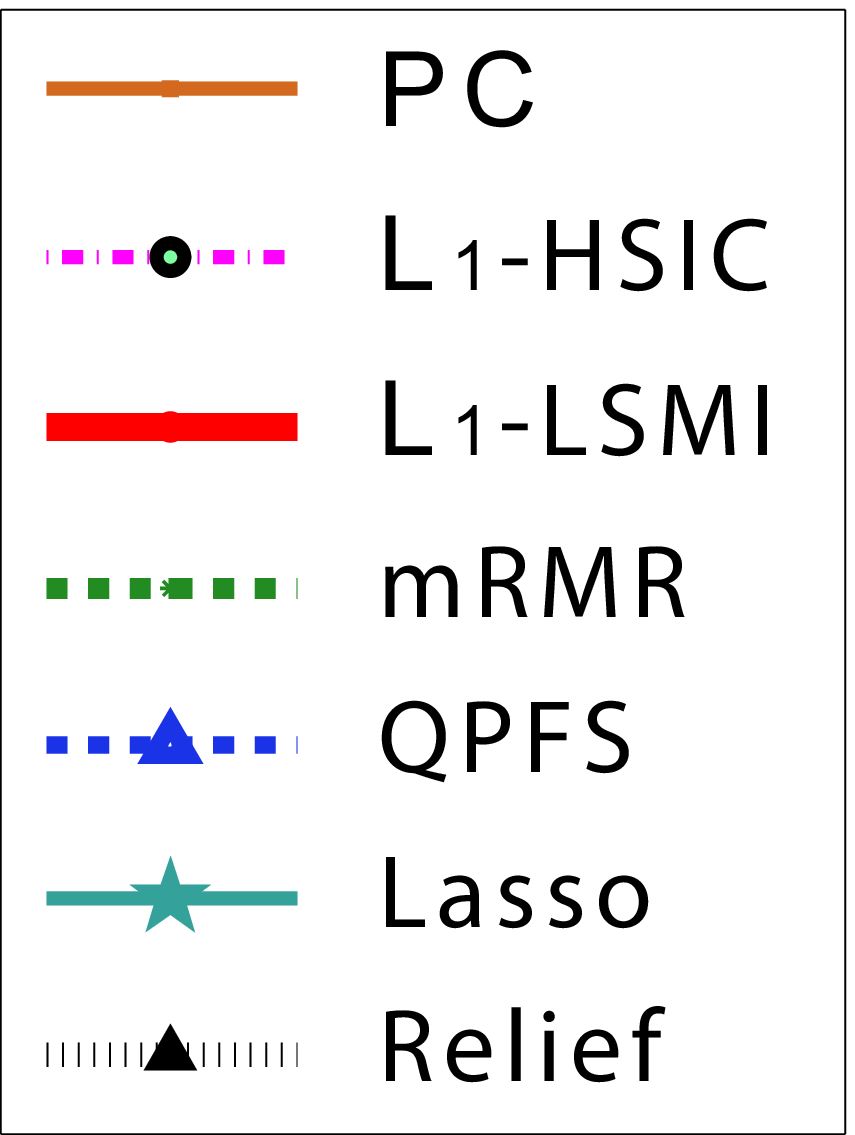}
~~~~~~~~~~~~~~~~~~~~~
}
\caption{Comparison of SVC/SVR errors of features selected by PC, $\ell_1$-HSIC, \proposed{}, mRMR, QPFS, Lasso and Relief.}
\label{fig:real_exp_noseq}
\end{figure*}

The experiment is repeated 20 times with $n=400$ points sampled in each trial. 
In each trial, $k$ is varied in the low range with a step size proportional to 
the entire dimensionality $m$.
For classification, each selected $k$-feature subset is scored with the test error of
a support vector classifier (SVC) with Gaussian kernels. 
For regression, the root mean squared error of support vector regression (SVR)
with Gaussian kernels is used.
The hyper-parameters of SVC and SVR are chosen with cross validation.
We use the implementations of SVC and SVR given in 
the LIBSVM library \cite{Chang2001}\footnote{LIBSVM: \url{http://www.csie.ntu.edu.tw/~cjlin/libsvm/}}. The results are shown in 
\figref{fig:real_exp_noseq}.

Overall, results suggest that using LSMI can give better features than HSIC (judged by the error
of SVC/SVR). This shows the importance of the availability of a model selection
criterion. 
\proposed{} and mRMR are competitive, especially on multi-class classification problems with many classes (e.g., segment and satimage). 
This is in contrast to PC and Relief which do not handle multi-class problems well.
As in the case of the toy data experiment, PC does not perform well in most cases since it does not take redundancy among features into account. 
An exception would be the \texttt{senseval2} problem in which PC performs the best among others. 
This is because 50 features in the \texttt{senseval2} dataset are derived from the first 50
principal components obtained by principal component analysis. 
Since principal components are orthogonal by definition, no redundancy has to be considered for this problem.
In some cases, considering feature redundancy may hurt the performance. 
This can be seen on \texttt{image}, \texttt{cpuact}, \texttt{senseval2}, and \texttt{musk2}
datasets when PC outperforms QPFS, suggesting that
features may not be correlated. Thus, ignoring redundancy and considering just relevancy gives a better performance. 
$\ell_1$-HSIC performs well in many cases,
but the performance may become unstable when $k$ is high due to the mentioned fact that 
$\sigma_{\mathrm{med}}$ also gets larger.

To objectively compare the performance, another experiment with the same setting is carried out on 22 datasets. 
The number of trials is set to 50.
For each method and dataset, $k$ is set to either 4, 10, or 20 depending on how large $m$ is. 
The selected $k$-feature subsets are evaluated by SVC or SVR, as in the previous experiment. 
The results are given in \tabref{tab:real_small}, where for each dataset, the method with the best
performance is shown in bold face. 
Other methods which have insignificant performance difference (based on
the one-sided paired t-test with 5\% significance level) to the best one are also marked
in the same way. Note that Lasso works on only binary and regression problems. 
Thus, the results for multi-class problems are not available. 
For F-HSIC and F-LSMI, we omit the results on the \texttt{ctslices} and \texttt{isolet}
datasets due to the
large computation time involved.

From the table, it can be seen quantitatively that overall \proposed{} performs
the best by judging from the number of times it ranks top.
Interestingly, although worse on small datasets, the performance of mRMR
approaches that of \proposed{} on high-dimensional datasets (i.e., the \texttt{musk1},
\texttt{musk2}, \texttt{ctslices}, and \texttt{isolet} datasets).
One reasonable explanation for this phenomenon is that, a
large number of features provide more freedom in choosing an alternative subset.
Even though there are interacting features, there may be many other alternative
non-interacting subsets which give an almost equivalent explanatory power. 
For this reason, the fact that mRMR cannot detect interacting features may be less significant.

\begin{table*}[t]
\caption{SVC/SVR errors of the features selected by PC, F-HSIC, F-LSMI, $\ell_1$-HSIC, \proposed{}, mRMR, QPFS, Lasso, and Relief on real
datasets.}
\label{tab:real_small}
\centering
{\scriptsize
\begin{tabular}{@{}l|c@{\ \ }c@{\ \ }c|lllllllll@{}}
\hline 
Dataset & $m$ & $n$ & $k$ & PC & F-HSIC & F-LSMI & $\ell_1$-HSIC & $\ell_1$-LSMI & mRMR & QPFS & Lasso & Relief \\ \hline \hline 
abalone (R) & 8 & 400 & 4 & 0.73 (.04) & 0.74 (.04) & 0.70 (.05) & 0.73 (.04) & 0.70 (.05) & 0.73 (.05) & 0.75 (.04) & 0.70 (.04) & \textbf{0.69 (.04)} \\ 
bcancer (C2) & 9 & 277 & 4 & 0.24 (.00) & 0.24 (.00) & 0.23 (.01) & \textbf{0.23 (.00)} & \textbf{0.23 (.01)} & 0.25 (.00) & 0.23 (.00) & 0.24 (.00) & 0.26 (.00) \\ 
glass (C6) & 9 & 214 & 4 & 0.29 (.00) & \textbf{0.28 (.00)} & 0.30 (.01) & 0.30 (.01) & 0.30 (.01) & 0.30 (.00) & 0.29 (.00) & -- & 0.31 (.00) \\ 
housing (R) & 13 & 400 & 4 & 4.03 (.19) & 4.14 (.20) & 4.20 (.21) & 3.95 (.20) & \textbf{3.91 (.19)} & 3.97 (.20) & 4.11 (.23) & 4.14 (.27) & 4.10 (.21) \\ 
vowel (C11) & 13 & 400 & 4 & \textbf{0.20 (.02)} & 0.23 (.03) & 0.24 (.03) & 0.20 (.02) & 0.21 (.02) & \textbf{0.20 (.02)} & 0.20 (.02) & -- & 0.21 (.02) \\ 
wine (C3) & 13 & 178 & 4 & 0.03 (.00) & 0.03 (.00) & \textbf{0.03 (.01)} & 0.03 (.01) & \textbf{0.03 (.01)} & 0.03 (.00) & 0.03 (.00) & -- & 0.03 (.00) \\ 
image (C2) & 18 & 400 & 4 & 0.10 (.01) & 0.19 (.03) & 0.17 (.03) & 0.13 (.03) & 0.06 (.02) & 0.14 (.02) & 0.11 (.02) & 0.11 (.02) & \textbf{0.05 (.01)} \\ 
segment (C7) & 18 & 400 & 4 & 0.19 (.03) & 0.24 (.03) & 0.17 (.02) & 0.11 (.03) & \textbf{0.05 (.01)} & \textbf{0.05 (.01)} & 0.08 (.03) & -- & 0.13 (.02) \\ 
vehicle (C4) & 18 & 400 & 4 & 0.32 (.02) & 0.33 (.03) & \textbf{0.28 (.02)} & 0.34 (.03) & \textbf{0.27 (.02)} & 0.39 (.05) & 0.39 (.05) & -- & 0.32 (.04) \\ 
german (C2) & 20 & 400 & 4 & \textbf{0.25 (.02)} & 0.29 (.01) & 0.29 (.02) & 0.25 (.02) & \textbf{0.25 (.02)} & 0.25 (.02) & 0.25 (.02) & 0.25 (.02) & 0.26 (.02) \\ 
cpuact (R) & 21 & 400 & 4 & 0.25 (.03) & 0.33 (.12) & 0.28 (.07) & 0.54 (.31) & \textbf{0.25 (.16)} & \textbf{0.23 (.06)} & 0.27 (.04) & 0.26 (.04) & 0.37 (.09) \\ 
ionosphere (C2) & 33 & 351 & 4 & 0.07 (.00) & \textbf{0.07 (.00)} & 0.08 (.01) & 0.07 (.00) & 0.07 (.00) & 0.09 (.00) & 0.07 (.00) & 0.07 (.00) & 0.07 (.00) \\ \hline
satimage (C6) & 36 & 400 & 10 & 0.22 (.02) & 0.14 (.01) & \textbf{0.13 (.02)} & 0.14 (.02) & \textbf{0.13 (.02)} & 0.14 (.01) & 0.14 (.02) & -- & 0.16 (.02) \\ 
spectf (C2) & 44 & 267 & 10 & 0.19 (.00) & 0.17 (.00) & \textbf{0.17 (.01)} & 0.19 (.01) & \textbf{0.17 (.01)} & 0.18 (.00) & 0.18 (.00) & 0.18 (.00) & 0.18 (.00) \\ 
senseval2 (C3) & 50 & 400 & 10 & \textbf{0.18 (.01)} & 0.18 (.01) & 0.18 (.02) & 0.19 (.02) & \textbf{0.18 (.01)} & 0.18 (.01) & \textbf{0.18 (.01)} & -- & 0.21 (.01) \\ 
speech (C2) & 50 & 400 & 10 & 0.01 (.00) & 0.01 (.00) & 0.01 (.00) & 0.01 (.00) & 0.01 (.00) & 0.02 (.00) & 0.01 (.00) & \textbf{0.01 (.00)} & 0.03 (.00) \\ 
sonar (C2) & 60 & 400 & 10 & 0.23 (.00) & 0.22 (.00) & \textbf{0.14 (.02)} & 0.21 (.02) & 0.16 (.02) & 0.18 (.00) & 0.19 (.00) & 0.16 (.00) & 0.19 (.00) \\ 
msd (R) & 90 & 400 & 10 & 0.95 (.06) & 0.94 (.06) & \textbf{0.92 (.06)} & 0.94 (.06) & 0.93 (.06) & 0.97 (.06) & 0.94 (.06) & \textbf{0.92 (.06)} & 0.96 (.06) \\ \hline
musk1 (C2) & 166 & 400 & 20 & 0.19 (.02) & 0.17 (.02) & 0.14 (.02) & 0.16 (.02) & 0.16 (.02) & 0.15 (.02) & 0.18 (.02) & \textbf{0.13 (.01)} & 0.19 (.03) \\ 
musk2 (C2) & 166 & 400 & 20 & 0.09 (.01) & 0.08 (.01) & \textbf{0.07 (.01)} & 0.09 (.01) & 0.08 (.01) & 0.09 (.01) & 0.09 (.02) & 0.07 (.01) & 0.09 (.01) \\ 
ctslices (R) & 379 & 400 & 20 & 0.79 (.07) & -- & -- & 0.64 (.05) & 0.60 (.07) & 0.45 (.04) & 0.46 (.02) & \textbf{0.41 (.03)} & 0.56 (.05) \\ 
isolet (C26) & 617 & 400 & 20 & 0.54 (.03) & -- & -- & 0.36 (.04) & \textbf{0.27 (.03)} & 0.30 (.03) & 0.30 (.03) & -- & 0.49 (.03) \\ \hline 
\multicolumn{4}{r|}{Top Count} & 3 & 2& 7 & 1 & 11 & 3 & 1 & 4 & 2 \\  \hline 
\end{tabular}
}
\end{table*}


\section{Conclusion}
\label{sec:conclude}
Feature selection is an important dimensionality reduction technique which can
help improve the prediction performance and speed, and facilitate the interpretation of a learned
predictive model. There are a number of factors which cause
the difficulty of feature selection. These include non-linear
dependency, feature redundancy, and feature interaction.

The proposed \proposed{} is an $\ell_1$-based algorithm that maximizes SMI between the
selected feature and the output. The main idea is to learn a sparse
feature weight vector whose coefficients can be used to determine the importance
of features. Only features corresponding to the non-zero coefficients in the
weight vector need to be kept. The use of $\ell_1$-regularization
allows simultaneous consideration
of features, which is essential in detecting a group of interacting features.
By combining with SMI which is able to detect a non-linear dependency, and implicitly handle feature redundancy, a
powerful feature selection algorithm is obtained. 

Extensive experiments were conducted to confirm the usefulness of \proposed{}.
We therefore conclude that \proposed{} is a promising method for practical use.

\section*{Acknowledgments}
We thank Dr.~Makoto Yamada for his valuable comments.
WJ acknowledges the Okazaki Kaheita International Scholarship Foundation,
HH acknowledges the FIRST Program,
and
MS acknowledges the MEXT KAKENHI 23120004.

\bibliography{l1lsmi_journal}

\end{document}